\documentclass[journal]{IEEEtran}

\usepackage{algorithmic}
\usepackage{algorithm}
\usepackage[linesnumbered,ruled,algo2e]{algorithm2e}
\SetAlCapFnt{\footnotesize}
\SetAlCapNameFnt{\footnotesize}

\usepackage{array}
\usepackage{textcomp}
\usepackage{stfloats}
\usepackage{url}
\usepackage{verbatim}
\usepackage{graphicx}
\usepackage{cite}
\usepackage{booktabs} 
\usepackage{multicol}
\usepackage{xfrac}

\DeclareFontFamily{U}{BOONDOX-cal}{\skewchar\font=45}{\itshape}
\DeclareFontShape{U}{BOONDOX-cal}{m}{n}{
<-> s*[1.05] BOONDOX-r-cal}{}
\DeclareFontShape{U}{BOONDOX-cal}{b}{n}{
<-> s*[1.05] BOONDOX-b-cal}{}
\DeclareMathAlphabet{\mathcalboondox}{U}{BOONDOX-cal}{m}{n}
\SetMathAlphabet{\mathcalboondox}{bold}{U}{BOONDOX-cal}{b}{n}
\DeclareMathAlphabet{\mathbcalboondox}{U}{BOONDOX-cal}{b}{n}
\usepackage{xcolor}
\usepackage{amsmath,mathtools,amsfonts} 
\usepackage[]{todonotes}
\usepackage{multirow} 
\usepackage[caption=false,font=footnotesize]{subfig} 
\usepackage{hyperref}
\hypersetup{
colorlinks,
citecolor=black,
filecolor=black,
linkcolor=black,
urlcolor=black
}

\newcommand{\norm}[1]{\left\lVert#1\right\rVert}

\newcommand{\refframe}[1]{\{#1\}}
\newcommand{\rotframe}[1]{\left[#1\right]}

\newcommand{\T}[3]
{\prescript{#1}{#3}{\boldsymbol{T}}_{#2}}

\newcommand{\R}[3]
{\prescript{#1}{#3}{\boldsymbol{R}}_{#2}}

\newcommand{\e}[3]
{\prescript{#1}{#2}{\boldsymbol{e}}_{#3}}

\newcommand{\p}[3]
{\prescript{#1}{#3}{\boldsymbol{p}}_{#2}}

\newcommand{\Tdot}[3]
{\prescript{#1}{#3}{\vphantom{T}}\dot{\boldsymbol{T}}_{#2}}

\newcommand{\Rdot}[3]
{\prescript{#1}{#3}{\vphantom{R}}\dot{\boldsymbol{R}}_{#2}}

\newcommand{\twist}[4]
{\vphantom{\textbf{\textsf{t}}}_{#2}^{#3}\boldsymbol{\mathcalboondox{t}}_{#4}^{#1}\vphantom{\textbf{\textsf{t}}}}

\newcommand{\wrench}[4]
{\vphantom{\textbf{\textsf{t}}}_{#2}^{#3}\boldsymbol{\mathcalboondox{w}}_{#4}^{#1}\vphantom{\textbf{\textsf{t}}}}

\newcommand{\zerom}[1]{\boldsymbol{\mathit{0}}_{\hspace*{-0.4ex}#1}}

\newcommand{\vect}[1]{\boldsymbol{#1}} 
\newcommand{\mat}[1]{\boldsymbol{#1}} 
\newcommand{\screw}[1]{\boldsymbol{\mathcalboondox{#1}}} 
\newcommand{\screwmv}[1]{\boldsymbol{\widetilde{\mathcalboondox{#1}}}} 

\begin{document}


\title{Automatic Derivation of an Optimal Task Frame for Learning and Controlling Contact-Rich Tasks}

\author{Ali~Mousavi~Mohammadi${}^{1}$, Maxim~Vochten${}^{1}$, Erwin~Aertbeli\"en${}^{1}$, Joris~De~Schutter${}^{1}$
\thanks{This result is part of a project that has received funding from the European Research Council (ERC) under the European Union's Horizon 2020 research and innovation programme (Grant agreement No. 788298).}
\thanks{${}^{1}$ Department of Mechanical Engineering, KU Leuven \& Core Lab ROB, Flanders Make, 3001 Leuven, Belgium.}}



\maketitle

\begin{abstract}

\textcolor{black}{
	In previous work on learning and controlling contact-rich tasks, the procedure for choosing a proper reference frame to express learned signals for the motion and the interaction wrench is often implicit, requires expert insight, or starts from proposed frame candidates.
	This article presents an automatic method to derive the optimal reference frame, referred to as optimal task frame, directly from the recorded motion and wrench data of the demonstration. 
	Using screw theory, several origin and orientation candidates are generated that maximize decoupling in the data. These candidates are then processed probabilistically, without needing hyperparameters, to obtain the optimal task frame. Its origin and orientation are independently fixed to either the world or the robot tool. The method works regardless of whether the task involves translation, rotation, force, or moment, or any combination thereof.
	The method was validated for various tasks, including surface following and manipulation of articulated objects, showing good agreement between derived and assumed expert task frames.
	To validate the robot's performance, a constraint-based controller was designed based on the data expressed in the derived task frames. These experiments demonstrated the approach's effectiveness and versatility.
	The automatic task frame derivation approach supports learning methods to design controllers for a wide range of contact-rich tasks.
}
\end{abstract}

\begin{IEEEkeywords}
Learning from demonstration, contact-rich task, task frame, screw theory, constraint-based control.
\end{IEEEkeywords}


\section{Introduction}


\IEEEPARstart{C}{ontact-rich} tasks in robotics are object manipulation tasks that involve contact with the environment or with other objects. In this paper, we focus on the manipulation of a single object attached to the robot end effector in contact with another object attached to the environment. Such tasks are abundant in all fields of robotics: industry, domestic robots, health care, etc. They include, for example, assembly, machining, picking and placing, and handling of articulated objects such as doors or drawers. 

Programming contact tasks on robot manipulators can be challenging since it requires coordinating and controlling both the desired motion of an object that is manipulated and its desired interaction wrench with the environment. \color{black} Depending on the task, the translational and/or the rotational motion may be relevant, as well as the interaction force and/or the interaction moment. \color{black}

\begin{figure}
\centering
\subfloat[]{
	\includegraphics[width=1.0\linewidth]{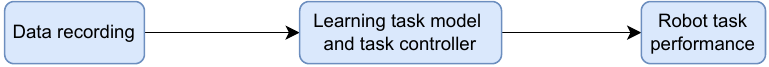}
	\label{fig:overview_a}
} \\ 
\subfloat[]{
	\includegraphics[width=1.0\linewidth]{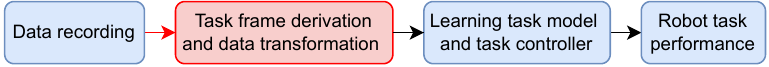}
	\label{fig:overview_b}
}
\caption{\textcolor{black}{(a) The typical workflow in robot task learning consists of data recording, learning a task model and controller from the data, and performing the task on a robot. (b) Our approach adds the automatic derivation of an optimal task frame from the data, prior to learning the task model.}}
\label{fig:overview}
\end{figure}

\color{black}
Learning from Demonstration (LfD) can help to make the programming of contact tasks more intuitive since it allows non-expert users to teach tasks to robots without a deep knowledge of robotics \cite{ravichandar2020recent}. Fig.~\ref{fig:overview_a} illustrates a typical LfD workflow consisting of first recording the data of human demonstrations of the task, then learning a task model from the data and designing a corresponding controller, and finally using this controller on the robot to perform the task while adapting to new circumstances. 

For object manipulation tasks in contact, it is important to define the task controller in a suitable reference frame. We refer to such a frame as a \textit{task frame}. The task frame is defined as a coordinate frame, with an origin and an orientation, in which the motion and wrench signals are controlled in different directions towards desired values \cite{bruyninckx1996specification}. Given a suitable choice for the task frame, some of these signals become constant or zero, facilitating the design of a stable and robust controller. Therefore, the choice of the task frame is crucial for the success of the control strategy.

\subsection{Problem and Research Objective}

It is not always clear which choice of task frame is best for the controller.
%
In some learning methods, the task frame is chosen implicitly to be the same as the reference frame in which the data were recorded. However, this usually does not lead to an optimal choice. Alternatively, the task frame can be chosen based on the expert user's intuition for the application at hand. For some tasks, the frame is best defined with respect to the robot's tool. Other times, it is better defined with respect to the environment. Considering the orientation, sometimes the axes of the frame are best chosen in the main direction of the translation or the interaction force, and other times they should be in the directions of rotation or interaction moment. 

Some LfD approaches have aimed to assist the user by proposing several candidates for this reference frame, and using the demonstration data to decide which candidate is most suitable \cite{ureche2015task,kober2015learning,niekum2015learning,manschitz2018mixture,huang2018generalized}. However, this still requires defining such candidates beforehand. 

A more user-friendly approach would be to derive a suitable task frame directly from the demonstration data itself. To the best of the authors' knowledge, this has not been explored yet.
Most closely related are methods that learn the main directions of motion and force (or compliance) from the data in a given reference frame such as \cite{suomalainen2016learning,suomalainen2017geometric,suomalainen2018learning,li2022learning,li2023augmentation}.
These main directions could be used to define the orientation of the task frame, as shown in \cite{conkey2019learning} for force directions. However, since the directions are learned in a predefined reference frame, they are fixed to this reference frame, for example, the world or tool frame. 
Furthermore, deriving an optimal origin for the controller from data, i.e., the origin of the task frame, does not seem to have been considered yet in the literature.
The objective of this paper is to propose a general method to automatically derive an optimal origin and orientation of the task frame from recorded motion and wrench data. To enable non-expert users to more easily learn a large variety of contact tasks, the method needs to have no restriction on the number of degrees of freedom in the task, and no prior assumption on whether the task is dominant in translation or rotation, or whether the task is dominant in force or moment.

\color{black}

\subsection{Approach}

\textcolor{black}{Fig.~\ref{fig:overview_b} highlights the step in the LfD workflow that we aim to optimize and automate.} To derive the task frame, we first establish two principles for an optimal choice of the task frame \textcolor{black}{based on maximizing the decoupling of signals in the controller}. Based on these principles we \textcolor{black}{generate multiple candidates for the origin and the orientation of the task frame from the demonstrated motion and wrench data, assuming that the data are segmented. We also assume that these candidates are either fixed to the world or to the tool. These candidates are processed to obtain the optimal task frame origin and orientation.} Once the task frame has been derived, the desired motion and wrench signals are expressed in this frame and a controller can be designed to perform the task.

The automatic task frame derivation procedure is validated for a diverse set of contact-rich tasks, without any assumption on the type of task, by comparing the result with an expert's choice. The tasks are performed on the robot after designing a constraint-based controller based on the derived task frame. The controller's robustness and generalization capabilities are tested under disturbances, parameter variations, and alterations in the task frame.

\subsection{Contribution and Outline}

The main contribution of this research is a multi-step algorithm to derive an optimal task frame from demonstrated motion and wrench data. \color{black} The main discerning features of our approach are that it incorporates data not only of the object's position and the interaction force, but also of the object's orientation and the interaction moment. \color{black} It decides the most appropriate reference frame separately for the origin and orientation of the task. Finally, it decides the most appropriate variable in which the motion and force signals should be parameterized. \textcolor{black}{The method is intended as a tool to automate the selection of a suitable reference frame before learning the task model with some learning method. The intent is to assist users that have limited experience in control.} 

The remainder of this article is organized as follows. Section~\ref{sec:related_works} \textcolor{black}{provides a more in-depth review of the literature on contact task learning related to the work proposed in this paper.} Section~\ref{sec:preliminaries} presents mathematical preliminaries, while Section~\ref{sec:task_frame_derivation} proposes the methodology to automatically derive the optimal task frame from motion and wrench data. Sections~\ref{sec:experiments} and \ref{sec:exp_results} \textcolor{black}{provide an extensive experimental validation of the task frame derivation procedure itself and of the use of the obtained task frame in the performance of learned tasks.} Finally, Section~\ref{sec:conclusion} outlines the major contributions, limitations, and conclusions of the work.

\color{black}\section{Relation to the state of the art}\color{black}
\label{sec:related_works}

\subsection{Collection of Demonstration Data}
Four common ways to collect demonstration data for LfD methods are kinesthetic teaching \cite{li2022learning,kim2022skill,zhao2022hybrid,conkey2019learning,suomalainen2016learning,ureche2015task,huang2018generalized}, teleoperation \cite{perez2017c,suomalainen2018learning}, passive observation \cite{peternel2017method,perico2020learning} and simulation. The first method involves physically grasping the robot and moving it through the task, recording motion via the robot's encoders and interaction wrench via torque sensors at the joints or via a force/torque sensor at the wrist. While cost-effective and fast, this method can be challenging for heavy robots \cite{suomalainen2018learning}, bulky robots \cite{el2019cobot}, redundant robots, dual-arm setups \cite{suomalainen2019improving}, or non-manipulator robots \cite{ravichandar2020recent}, as undesirable dynamics induced by the operator may affect the recorded data \cite{peternel2017method}. The second method is usually more expensive since it involves an external interface between the robot and the operator, such as a joystick, haptic device or graphical user interface. Collecting demonstration data by teleoperation is also more challenging than kinesthetic teaching \cite{fischer2016comparison}. In the third method, the operator uses an instrumented tool to perform the task and record motion and wrench data. In the last method, simulation environments like MoJuCo \cite{mujoco} and Bullet \cite{bullet} are used to generate demonstration data. The latter method offers abundant data which can enable reinforcement-learning-based methods to widely explore the task without any risk for the robot. However, transferring learned skills from the simulator to the real world is still challenging \cite{suomalainen2022survey}. Consequently, building a reliable model derived from one \cite{li2022learning,kim2022skill,li2023augmentation} or a few \cite{perico2020learning,zhao2022hybrid,ureche2015task} demonstrations is a critical aspect of learning contact-rich tasks. In this article, passive observation is employed to record demonstrations as it aligns with the most natural way for the operator to perform tasks \cite{el2019cobot}, but the developed approach is not limited to this demonstration method. \color{black} Furthermore, the proposed approach to derive an optimal task frame can also be used with simulation data as input or with robot data, i.e., motion and wrench data recorded during a task performed by a robot. \color{black}

\subsection{Assumptions and Applications}
Some researchers included methods for segmentation in their proposed pipelines for processing the demonstration data, such as \cite{ureche2015task,peternel2017method,li2022learning,li2023augmentation}, but we left segmentation out of the scope of the present paper. I.e., we assume demonstration data has already been segmented and propose an approach to \color{black} derive a task frame \color{black} for a single task segment in which the contact situation remains the same over time. On the other hand, while a lot of approaches in literature focus on LfD for a single task or a limited class of tasks, our aim was to develop a generic approach, without further knowledge or assumptions about the task other than that the contact is between only two objects with a measurable relative motion \color{black} and interaction wrench. \color{black}

\subsection{Selection of Task Frame}
\label{subsec: selection_tf}
To control a robot while manipulating an object, typically, a proper reference frame is chosen. This reference frame has different names in the literature such as \textit{compliance frame} \cite{mason1981compliance}, \textit{constraint frame} \cite{conkey2019learning}, and \textit{task frame} \cite{bruyninckx1996specification} but the main idea behind it remains the same.
It was first proposed by Mason \cite{mason1981compliance}, where a task was specified in terms of natural (i.e. physical) and artificial (i.e. imposed) constraints for motion and wrench in the context of frictionless contact between rigid bodies. Based on this idea, Raibert and Craig \cite{raibert1981hybrid} formulated a hybrid controller to obtain decoupling between the position- and force-controlled directions. The task frame is selected to simplify the specification of the natural and artificial constraints. Since this process requires a good understanding of the task geometry, it is entrusted to the expert \cite{raibert1981hybrid}.
De Schutter and Van Brussel \cite{de1988compliantI,de1988compliantII} introduced an adaptive or \textit{moving} task frame for task geometries such as in {2-D} contour following, where the directions of contact force and translational velocity are not fixed w.r.t. both world and tool, but are estimated from the motion and wrench measurements during the task performance.
Bruyninckx et al. \cite{bruyninckx1995kinematic} presented a generic task geometry involving contact between two objects with doubly curved surfaces and leading to a contact frame that moves w.r.t. \textit{both} objects in contact. This moving frame simplifies for special cases such as contact between polyhedral objects.
Bruyninckx and De Schutter provided more examples of task frame selections, including adaptive and non-adaptive ones \cite{bruyninckx1996specification}. They suggested to maximize the decoupling between motion and wrench directions, but did not present a detailed method for achieving this goal.

Task frame selection regained interest in the context of LfD, since demonstration data and task models have to be represented in an appropriate reference frame. There are two common ways to choose a task frame.
The first way is, still as in earlier works, to rely on implicit high-level robotics expertise, i.e., without using any prescribed method.
The second way is to select between several task frame candidates that are predefined based on expert knowledge. This selection is based on criteria like consistency \cite{ureche2015task}, variance \cite{kober2015learning} or generalization capability of the demonstration data \cite{manschitz2018mixture}. 
A first set of such studies \textit{directly} select between different task frame candidates \cite{ureche2015task,kober2015learning,niekum2015learning} or optimize the position and orientation of such 
candidates \cite{huang2018generalized}.
Other studies \textit{indirectly} learn an appropriate reference frame or components thereof. \color{black} For example, a method was proposed which learns to activate a set of attractors that are each defined in a different appropriate reference frame \cite{manschitz2018mixture}. In \cite{conkey2019learning}, the authors proposed a method that could adapt the orientation of the task frame w.r.t. the environment. Additionally, some works learned the main directions of motion and force (or compliance) from the data in a given reference frame, such as \cite{suomalainen2016learning,suomalainen2017geometric,suomalainen2018learning,li2022learning,li2023augmentation}. \color{black}

In this paper, we propose a \textit{generic} approach to derive an optimal task frame for contact-rich tasks, allowing interaction with both articulated and non-articulated objects. Both the origin and the orientation of the task frame are derived. The origin and orientation can be fixed in different viewpoints (world or tool) so that the task frame as a whole is not necessarily fixed to either world or tool. We use both motion and wrench data, consider both their translational and rotational vector components and automatically derive the motion and wrench \textit{vector components of interest} from the demonstration data. \color{black} The procedure does not involve hyperparameters. \color{black} While all works mentioned above exhibit some of these features, they lack several others. \color{black} In particular, to the best of the authors' knowledge, there is no method available in the literature to derive the task frame from motion and wrench data. Others use application-specific knowledge \cite{denivsa2015learning,suomalainen2016learning,suomalainen2017geometric,peternel2017method,perez2017c,abu2018force,suomalainen2018learning,qin2019robotic,conkey2019learning,perico2020learning,suomalainen2021imitation,zhao2022hybrid,li2022learning,li2023augmentation} or predefined candidates \cite{niekum2015learning,kober2015learning,ureche2015task,huang2018generalized,manschitz2018mixture}. Some works do not consider tasks involving articulated objects \cite{zhao2022hybrid,perico2020learning,conkey2019learning,qin2019robotic,suomalainen2018learning,manschitz2018mixture,huang2018generalized,peternel2017method,suomalainen2017geometric,suomalainen2016learning,ureche2015task,kober2015learning,niekum2015learning,suomalainen2021imitation} or non-articulated objects \cite{li2023augmentation,li2022learning}. Some works do not consider tasks with dominant rotation \cite{ureche2015task,suomalainen2016learning,suomalainen2017geometric,peternel2017method,perez2017c,abu2018force,manschitz2018mixture,suomalainen2018learning,qin2019robotic,conkey2019learning,zhao2022hybrid,li2022learning,li2023augmentation}, force \cite{niekum2015learning,huang2018generalized,manschitz2018mixture} or moment \cite{niekum2015learning,ureche2015task,racca2016learning,suomalainen2016learning,suomalainen2017geometric,peternel2017method,perez2017c,manschitz2018mixture,suomalainen2018learning,abu2018force,huang2018generalized,qin2019robotic,conkey2019learning,zhao2022hybrid,li2022learning,li2023augmentation}. \color{black} On the other hand, although it is not common to consider a \textit{moving} task frame for contact-rich tasks in the context of LfD, this was done in \cite{conkey2019learning}.


\subsection{Task Model and Control}
Typically, for learning contact-rich tasks from human demonstration, motion and wrench data are recorded to build a task model. However, in \cite{peternel2017method} muscle activity (EMG signals) was recorded as well to estimate the desired stiffness behavior of the robot end-effector.
Various methods have been explored to model skills required for contact-rich tasks.
These models should be able to reproduce the task under similar and novel situations. Some of the most common methods are Gaussian mixture model/regression \cite{abu2018force,zhao2022hybrid} and dynamic movement primitives \cite{peternel2017method,qin2019robotic,conkey2019learning,denivsa2015learning}.
From the learned models the desired values for motion and interaction wrench are generated. These signals need to be tracked by a controller.
Unlike position/force control, as applied in \cite{conkey2019learning}, impedance control is a popular solution for contact-rich tasks, with different types of controllers employed in
\cite{ureche2015task,denivsa2015learning,racca2016learning,suomalainen2017geometric,peternel2017method,abu2018force,qin2019robotic,zhao2022hybrid}. However, impedance behaviors during contact can also be achieved using an indirect formulation, as shown in \cite{perico2020learning}. Finally, the presence of hard geometric constraints is searched for in the demonstration data in \cite{li2022learning,li2023augmentation}, while such constraints are explicitly defined in the task model in \cite{perez2017c}.

The task model \color{black} used in the experimental validation Sections \ref{sec:experiments} and \ref{sec:exp_results} \color{black} is derived from 6-D motion and 6-D wrench data. It consists of 1) a task frame, with its location and orientation as parameters, and 2) the demonstration data, expressed in the task frame and averaged over the demonstration trials to find the desired values for the motion and interaction wrench. No further processing of the data is performed. A \textit{generic} constraint-based controller is \color{black} used \color{black} to follow these desired values. However, we believe that other learning methods could leverage this automatically derived task frame, for building their models and exploring alternative controllers.

\subsection{Generalization and Skill Transfer}
The learned model is expected to be able to perform the task not only under conditions similar to the demonstrations but also in novel scenarios. This challenge is commonly referred to as generalization. Recent studies explored aspects such as starting from new positions \cite{ureche2015task,kober2015learning,suomalainen2016learning,qin2019robotic}, reaching to new positions \cite{ureche2015task,kober2015learning}, positional displacements \cite{racca2016learning,li2022learning}, dealing with unexpected end-effector collisions \cite{zhao2022hybrid,conkey2019learning}, variations in contact friction \cite{ureche2015task,conkey2019learning}, and achieving faster performance of the learned task \cite{perico2020learning}.
Transferring learned skills to new objects and geometries is another crucial aspect. The model should be able to perform the task with similar but different objects, as discussed in \cite{ureche2015task,kober2015learning,perez2017c,peternel2017method,perico2020learning,li2022learning,kim2022framework,kim2022skill}. This involves preserving the key features of the demonstrated task while achieving the desired task goals.
In this paper, we also show how the proposed approach can be used to generalize acquired skills to novel scenarios and to transfer these skills to new objects using the parameters of the learned task model.

\section{Preliminaries and Notation}
\label{sec:preliminaries}

\subsection{Orientation and Pose}
\label{sec:prelim_tf}

In this article we distinguish between two types of reference frames that can be attached to a rigid body $\mathcal{B}$. The first type is referred to as an \textit{orientation frame} since only the orientation of this frame is relevant, not the origin. A general orientation frame $\rotframe{b}$ can be represented with respect to another frame $\rotframe{a}$ by a rotation matrix $\R{b}{a}{}$ $\in \mathbb{R}^{3\times3}$:
\begin{equation}
	\R{b}{a}{} =
	\begin{bmatrix}
		{}^{b}\vect{x}_a & {}^{b}\vect{y}_a & {}^{b}\vect{z}_a
	\end{bmatrix},
	\label{eq:rotmat}
\end{equation}
where the columns ${}^{b}\vect{x}_a$, ${}^{b}\vect{y}_a$, and ${}^{b}\vect{z}_a$ are the coordinates of the three orthogonal basis vectors of $\rotframe{b}$ expressed in $\rotframe{a}$.

Equation \eqref{eq:rotmat} represents the orientation of $\rotframe{b}$ with respect to $\rotframe{a}$ using coordinates expressed in $\rotframe{a}$. To analyze the same relative rotation, but using coordinates expressed in a third frame $\rotframe{c}$, a \textit{similarity transformation} can be used \cite{spong2020robot}:
\begin{equation}
	\label{eq:sim_transf1}
	\R{b}{a}{c} = \R{c}{a}{}^{-1} ~ \R{b}{a}{} ~ \R{c}{a}{}.
\end{equation}
The three successive rotation matrices are interpreted as follows: change the reference frame from $\rotframe{c}$ to $\rotframe{a}$ using $\R{c}{a}{}^{-1}$, then apply the relative rotation $\R{b}{a}{}$ in frame $\rotframe{a}$, and finally change the reference frame back to $\rotframe{c}$ using $\R{c}{a}{}$.
In robotics such a similarity transformation is also used to transform the coordinate system of an inertia matrix or screw matrix, which is better known as an \textit{adjoint transformation} or an \textit{adjoint map} \cite{murray1994,lynch2017modern}. 

The orientation of a frame can also be represented by a \textit{rotation vector} $\vect{r} \in \mathbb{R}^{3\times1}$. This rotation vector has the meaning of a rotation about an axis along the unit vector $\vect{r}/\norm{\vect{r}}$ by an angle $\norm{\vect{r}}$. The rotation matrix and rotation vector are related through the matrix exponential: 
\begin{equation}
	\mat{R} = \exp \left( [\vect{r}]_\times \right),
\end{equation}
where $\left[\cdot\right]_\times$ is an operator that produces the skew-symmetric cross product matrix from a vector. The inverse relation is established by the matrix logarithm:
\begin{equation}
	\label{eq:matrix_logarithm_R}
	\left(\log \mat{R}\right)^{\vee} = \vect{r}, 
\end{equation} 
where the operator $(\cdot)^{\vee}$ is the inverse of $\left[\cdot\right]_\times$ such that $\vect{r} = ([\vect{r}]_\times)^\vee$.

The second type of reference frame is referred to as a \textit{pose frame}. Both its orientation and its origin are relevant. The orientation of a pose frame $\refframe{b}$ can again be represented by the rotation matrix $\R{b}{a}{}$, while the position of its origin with respect to $\refframe{a}$ is given by a position vector ${}^{b}\vect{p}_{a} \in \mathbb{R}^{3\times1}$. The orientation and position of the pose frame can be combined in a $4\times4$ \textit{homogeneous transformation matrix} or \textit{pose matrix}:
\begin{equation}
	\T{b}{a}{} =
	\begin{bmatrix}
		\R{b}{a}{} & {}^{b}\vect{p}_{a} \\
		\zerom{1\times3}  & 1
	\end{bmatrix}.
\end{equation}

\begin{figure}[!t]
	\centering
	\includegraphics[width = 6cm, page = 1]{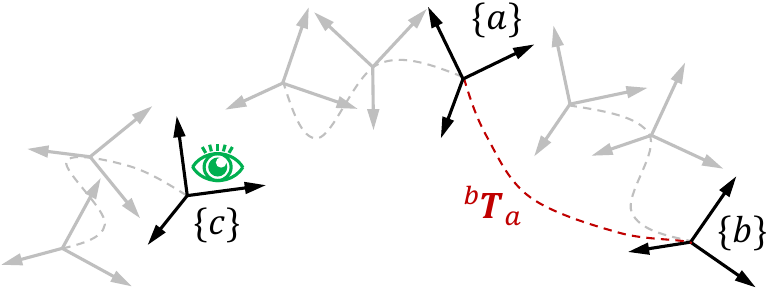}
	\caption{The relative pose $\T{b}{a}{}$ between frames $\refframe{a}$ and $\refframe{b}$ can be observed from the perspective of a third frame $\refframe{c}$. Without loss of generality, any of these frames could be stationary.}
	\label{fig:sim_transf}
\end{figure}

To analyze the relative pose between $\refframe{b}$ and $\refframe{a}$ using coordinates expressed in a third frame $\refframe{c}$, we extend the \textit{similarity transformation} of \eqref{eq:sim_transf1} to pose matrices:
\begin{equation}
	\label{eq:sim_transf}
	\T{b}{a}{c} = \T{c}{a}{}^{-1} ~ \T{b}{a}{} ~ \T{c}{a}{}.
\end{equation}
Fig.~\ref{fig:sim_transf} visualizes the similarity transformation under general conditions with three moving frames.

The pose of a frame can also be represented by a \textit{displacement screw} $\screw{d} =( 
\vect{r}^T ~~ \vect{u}^{o\,T}
)^T\in\mathbb{R}^{6\times1}$ , where $\vect{r}$ is again the rotation vector and $\vect{u}^o$ is the translation at the origin of the reference frame. They are related by the matrix exponential:
\begin{equation}
	\mat{T} = \exp \left( [\screw{d}]_\times \right),
\end{equation}
where $\left[\cdot\right]_\times$ is an operator that produces the following matrix from a screw: 
\begin{equation}
	\left[ \screw{d} \right]_\times = \begin{bmatrix}
		\left[\vect{r}\right]_\times & \vect{u}^o \\ \boldsymbol{0}_{1\times3} & 0
	\end{bmatrix}.
\end{equation}
The inverse relation is established by the matrix logarithm:
\begin{equation}
	\label{eq:displacement_screw}
	\left(\log \mat{T}\right)^{\vee} = \screw{d}, 
\end{equation} 
where the operator $(\cdot)^{\vee}$ applied to a screw is defined as the inverse of $\left[\cdot\right]_\times$ such that $\screw{d} = ([\screw{d}]_\times)^\vee$.

\subsection{Velocity kinematics}
\label{sec:prelim_screw}


The velocity kinematics of an orientation frame are characterized by the rotational velocity vector $\vect{\omega}$. When the vector's coordinates are expressed in $\rotframe{a}$, the relation between ${}_a\vect{\omega}$ and the derivative of the rotation matrix $\Rdot{b}{a}{}$ is:
\begin{align}
	\Rdot{b}{a}{} = [{}_{a}\vect{\omega}]_{\times}  \R{b}{a}{}.  
\end{align}

The velocity kinematics of a pose frame are characterized by the \textit{twist}  $\screw{t} = ( 
\vect{\omega}^T ~~ \vect{v}^{o\,T}
)^{T} \in \mathbb{R}^{6\times1}$, consisting of the rotational velocity vector $\vect{\omega}$ and the translational velocity vector $\vect{v}^o$ at the origin $o$ of the reference frame. When the twist coordinates are expressed in $\refframe{a}$, the relation between ${}_{a}\screw{t}$ and the time-derivative of the pose matrix $\Tdot{b}{a}{}$ can be worked out as:
\begin{align}
	\label{eq:pose_differentiation}
	\Tdot{b}{a}{} = [{}_{a}\screw{t}]_{\times}  \T{b}{a}{}.  
\end{align}

\subsection{Wrench and general screw}

The interaction force and moment between two rigid bodies can be represented using a \textit{wrench} $\screw{w} = ( \vect{f}^T ~~ \vect{m}^{o\,T} )^{T} \in \mathbb{R}^{6\times1}$, where $\vect{f}$ and $\vect{m}^o$ are the interaction force vector and interaction moment vector at the origin $o$ of the reference frame respectively.

The twist and wrench can both be interpreted as \textit{screws} according to screw theory \cite{ball1900treatise}. For twists, the theorem of Mozzi-Chasles holds that the motion of a rigid body can always be represented as a rotation and translation along an axis in space, the \textit{screw axis for motion} \cite{murray1994,featherstone2014rigid}. Similarly, Poinsot's theorem states that all forces and moments acting on a rigid body can always be represented as a force and moment along an axis in space, the \textit{screw axis for wrench}. 

Both the screw twist and wrench will be represented in this paper with a \textit{general screw} $\screw{s} = ( \vect{a}^T ~~ \vect{b}^{o\,T} )^{T}$ where $\vect{a}$ is the directional component of the screw along the screw axis, $\vect{b}^o$ is the moment component of the screw, i.e. the moment about the screw axis seen at the origin of the reference frame, and the coordinates of vectors $\vect{a}$ and $\vect{b}^o$ are expressed in the reference frame.

A screw $\screw{s}$ that is expressed in reference frame $\refframe{b}$ can be transformed to $\refframe{a}$ using a \textit{screw transformation matrix} $\mat{S}$:
\begin{equation}
	\label{eq:screw_ref_change}
	{}_{a}\screw{s} =  \boldsymbol{S}\left(\T{b}{a}{}\right) ~ {}_{b}\screw{s},
\end{equation}
with:
\begin{equation}
	\label{eq:screw_transf}
	\boldsymbol{S}\left(\T{b}{a}{}\right) = 
	\begin{bmatrix}
		\R{b}{a}{} & \zerom{3\times3} \\
		\left[\p{b}{a}{}\right]_{\times} \R{b}{a}{} & \R{b}{a}{}
	\end{bmatrix}.
\end{equation}
Note that applying this transformation for screws bears a relation with the similarity transformation for poses \eqref{eq:sim_transf}, as also explained in \cite{murray1994,lynch2017modern}.

A special case of \eqref{eq:screw_ref_change} where only the application point of the moment component $\vect{b}^{o}$ needs to be changed from a point $\p{o_1}{}{}$ to $\p{o_2}{}{}$, can succinctly be written as:
\begin{equation}
	\label{eq:refpoint_transf}
	{}_{}\vect{b}^{o_2} = \left[\p{o_2}{}{} - \p{o_1}{}{} \right]_\times \vect{a} + {}_{}\vect{b}^{o_1},
\end{equation}
in which the coordinates of all vectors are expressed with respect to the same reference frame.

\subsection{Time-invariant motion and wrench trajectories}
\label{sec:progress}

\color{black}Motion and wrench trajectories are usually considered as a function of time $t$, such as the pose  $\T{}{}{}(t)$, twist $\screw{t}(t)$, and wrench $\screw{w}(t)$ trajectories. 
They can be made time-invariant by re-parameterizing them as a function of a geometric progress variable $\xi$. There are multiple valid choices for the progress variable. For example, it can be defined as the cumulative arc length of the trajectory traversed by a chosen reference point, or as the cumulative rotation angle, or by some combination of the two. The progress rate $\dot{\xi}$ would then correspond to the traversed arc length or rotation angle per unit time, or their combination. Often, the progress variable is also made non-dimensional (ranging from $0$ to $1$) so that trajectories of different length can be compared.

For a given progress $\xi(t)$ as a function of time, the reparameterizations of pose, wrench, and twist from the time domain to the progress domain are defined as follows \cite{vochten2023invariant}:
\begin{align}
	\T{}{}{}{}(\xi) = \T{}{}{}{}( t(\xi) ),	~ {}_{}^{}\screw{w}(\xi) = {}_{}^{}{\screw{w}}( t(\xi) ), ~ \screw{t}(\xi) = \frac{\screw{t}( t(\xi) )}{\dot{\xi}( t(\xi) )},
	\label{eq:reparam4}
\end{align}
where $t(\xi)$ is the inverse function of $\xi(t)$. 

We can go back from the progress domain to the time domain using:
\begin{align}
	\T{}{}{}{}(t) = \T{}{}{}{}( \xi(t) ),	~ {}_{}^{}\screw{w}(t) = {}_{}^{}{\screw{w}}( \xi(t) ), ~ \screw{t}(t) = \screw{t}( \xi(t) ){\dot{\xi}( t )}.
	\label{eq:reparam4-inv}
\end{align}
Consecutive application of \eqref{eq:reparam4} and \eqref{eq:reparam4-inv} with different progress timings $\xi(t)$ results in a different time trajectory $\T{}{}{}{}(t)$, i.e., the time-dependency of the trajectory has changed, but the spatial path has remained the same.

These re-parameterizations are helpful to master the variability in demonstrations by allowing us to make time-invariant models. \color{black}
In practice, for discrete-time data, these reparameterizations are performed using numerical interpolation.
Throughout this paper and in order not to overload notation, we do not explicitly mention the dependence of signals on time or on another progress variable when such dependence is obvious from the context.

\subsection{Average Vector Orientation Frame (AVOF)}
\label{sec:prelim_avof}

The \textit{Average Vector Orientation Frame} (AVOF) is an orientation frame that characterizes the average direction and main directions of variation for a given set of $N$ unordered vectors $\{\vect{c}_i\}$, with $i=1\dots N$. The concept was originally introduced in \cite{ancillao2022optimal} for characterizing the (variation in) orientation of the screw axis corresponding to a set of twists $\{\screw{t}_i\}$. Later it was extended in \cite{vochten2023invariant} to characterize sets of general vectors $\{\vect{c}_i\}$. 

The AVOF is calculated as follows. First construct the \textit{uncentered} covariance matrix from the vectors $\{\vect{c}_i\}$:
\begin{align}
	\label{eq:cov_avof}
	\mat{C}_{c} &= \frac{1}{N} \sum_{i=1}^N \vect{c}_i \vect{c}_i^T.
\end{align}
(The data is uncentered so that the average direction of the vectors is included.) Next, the \textit{Singular Value Decomposition} (SVD) of the covariance matrix $\mat{C}_{c}$ is determined. Since 
$\mat{C}_{c}$ is a symmetric matrix, the SVD can be simplified to:
\begin{align}
	\label{eq:svd_avof}
	\mat{C}_{c} = \mat{U}_{avof} \ \mat{\Sigma}_{avof} \ \mat{U}_{avof}^T \text{,}
\end{align}
where $\mat{U}_{avof}$ is a right-handed orthonormal matrix consisting of three singular vectors and $\mat{\Sigma}_{avof}$ is a diagonal matrix containing the singular values of $\mat{C}_{c}$: $\sigma_1^2 \geq \sigma_2^2 \geq \sigma_3^2$.  

\begin{figure}[!t]
	\centering
	\includegraphics[width = 5.5cm]{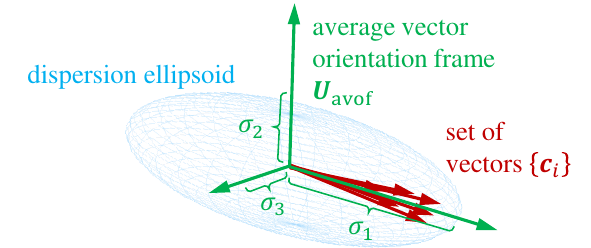}
	\caption{Average vector orientation frame for a set of vectors \color{black} $\{\vect{c}_i\}$ \color{black} together with a dispersion ellipsoid characterizing the main directions of variation.}
	\label{fig:avof}
\end{figure}

The AVOF's orientation corresponds to $\mat{U}_{avof}$. The first column of $\mat{U}_{avof}$ is interpreted as the average direction of all vectors $\{\vect{c}_i\}$. The second and third columns are the main directions of variation for the vectors. The overall magnitude of the vectors in these three directions is given by $\sigma_1$, $\sigma_2$, and $\sigma_3$, respectively. Fig.~\ref{fig:avof} provides an illustration of the AVOF where the spatial distribution of vectors is visualized using a dispersion ellipsoid corresponding to the SVD.

The difference between the AVOF method and other SVD-based approaches in the literature \cite{ureche2015task,suomalainen2016learning,conkey2019learning,li2022learning} is that in the AVOF method there is no explicit normalization of the vectors $\vect{c}_i$ before calculating the covariance matrix $\mat{C}_{c}$. The reason is that additive noise on a vector with a small magnitude significantly alters the direction of the normalized vector. To remove this sensitivity, the vectors  $\vect{c}_i$ are not normalized, so that in the SVD more weight is given to vectors with higher magnitudes than vectors with lower magnitudes. This can be interpreted as a weighted average of directions, where the weights correspond to the magnitudes. 

The variation on the estimated orientation of the AVOF, in terms of a covariance matrix  $\mat{C}_{avof}$, can be defined by scaling the covariance matrix $\mat{C}_{c}$ so that it becomes dimensionless. There are different options for scaling. Here, we divide $\mat{C}_{c}$ by the trace of $\mat{C}_{c}$, which is the same as the trace of $\mat{\Sigma}_{avof}$ and also the same as the average squared norm of the vectors $\vect{c}_i$:
\begin{equation}
	\label{eq:nondim1}
	\mat{C}_{avof} := \frac{\mat{C}_c}{\operatorname{trace}(\mat{C}_c)} = \frac{\mat{C}_c}{\operatorname{trace}(\mat{\Sigma}_{avof})} = \frac{\mat{C}_c}{\frac{1}{N} \sum_{i=1}^{N} \norm{ \vect{c}_{i}}^2}.
\end{equation}

The calculation of the AVOF orientation $\mat{U}_{avof}$ and the covariance matrix $\mat{C}_{avof}$ will be referred to by the function operator \texttt{avof} in the rest of this article. Algorithm \ref{alg:avof} summarizes the AVOF method.

\begin{algorithm2e}[t!]
	\footnotesize
	\caption{Average Vector Orientation Frame}
	\label{alg:avof}
	\SetKwFunction{AVOF}{avof}
	\SetKwProg{Fn}{Function}{:}{}
	\Fn{\AVOF{\color{black} $\{\vect{c}_i\}$\color{black}}}{
		\KwIn{ \color{black} $\{\vect{c}_i\}$ \color{black} with $i = 1 \dots N$ \tcp*{set of vectors}\\
		}
		\KwOut{%
			$\mat{U}_{avof}$, $\mat{C}_{avof}$ \; \\
		}
		$\mat{C}_{c}$ $\gets$ $\frac{1}{N} \sum_{i=1}^N \vect{c}_i \vect{c}_i^T$ \tcp*{covariance matrix} 
		$   \mat{U}_{avof}, \mat{\Sigma}_{avof}, \text{...}$ $\gets$ $\operatorname{svd}(\mat{C}_{c})$ \; 
		$ \mat{U}_{avof}(:,3) \gets \mat{U}_{avof}(:,1) \times \mat{U}_{avof}(:,2)$ \tcp*{right-handed} 
		$ \mat{C}_{avof}$ $\gets$  $\mat{C}_{c} / \operatorname{trace}(\mat{C}_c)$ \tcp*{make nondimensional}
	}
	\normalsize
\end{algorithm2e}

\subsection{\color{black}Average Screw-axes Intersection Point (ASIP)\color{black}}
\label{sec:prelim_asip}

The \color{black} \textit{Average Screw-axes Intersection Point} \color{black} (ASIP) is the approximate intersection point for a given set of $N$ unordered screw axes associated with screws $\{\screw{s}_i\}$ with $i = 1 \dots N$. In \cite{ancillao2022optimal}, a method was proposed to calculate the ASIP for a set of screw twists $\{\screw{t}_i\}$ by finding the point with minimal average velocity due to all screws. By considering the velocity instead of the distance to the screw axes, the resulting estimate of the ASIP was shown to be more robust to noise. This is because, in the presence of additive noise, the screw axis is not well-defined when the directional component $\vect{a}_i$ is very small. Later, the method was extended to general screws $\{\screw{s}_i\}$ in \cite{vochten2023invariant}, so that it could be applied to wrenches $\{\screw{w}_i\}$ in addition to twists $\{\screw{t}_i\}$. 

The ASIP for a set of screws $\{\screw{s}_i\}$ is defined as the point with \textit{minimal average moment} (i.e., translational velocity for $\{\screw{t}_i\}$ or moment for $\{\screw{w}_i\}$). This point can be calculated by solving the following least-squares problem:
\begin{equation}
	\p{}{asip}{} = \underset{\p{}{}{}}{\operatorname{argmin}} \frac{1}{N} \sum_{i=1}^{N} \left\| \vect{a}_i \times \p{}{}{} + \vect{b}^o_i \right\|^2.
	\label{eq:criterion1}
\end{equation}
This problem is ill-defined when all the directional vectors  $\{\vect{a}_i\}$ of the screws are parallel or close to zero. Therefore, a regularization term $\epsilon \left\| \p{}{}{} - \p{}{0}{} \right\|^2$ is added in the above criterion \eqref{eq:criterion1} with a weight $\epsilon$ that biases the solution towards a given prior value $\p{}{0}{}$. When $\epsilon$ is kept small, the effect is negligible in the regular case. The solution of the regularized least-squares problem can be worked out as:
\begin{align}
	\label{eq:criterion1_sol}
	\p{}{asip}{} &=  \left[ \mat{A}  + \epsilon  \mat{I}_3   \right]^{-1}
	\left(  \frac{1}{N} \sum_{i=1}^{N}  ( \vect{a}_i \times \vect{b}^o_i ) + \epsilon  \p{}{0}{}   \right),
\end{align}
with $\mat{A} = \frac{1}{N} \sum_{i=1}^{N}   [{\vect{a}}_i]_{\times} {[\vect{a}_i}]_{\times}^T $ and $\mat{I}_3$ a $3\times3$ identity matrix. 

\begin{figure}[t!]
	\centering
	\includegraphics[width = 6cm]{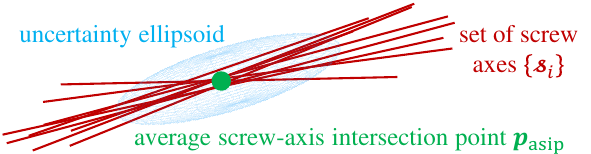}
	\caption{\color{black}Average Screw-axes intersection point $\p{}{asip}{}$ for a set of screw axes $\{\screw{s}_i\}$ and its corresponding uncertainty ellipsoid.\color{black}}
	\label{fig:asip}
\end{figure}

The variation on the estimate of the ASIP can be studied using the following covariance matrix \cite{ancillao2022optimal}:
\begin{equation}
	\label{eq:cov_asip}
	\mat{C}_{asip} = \operatorname{var}(\p{}{asip}{}) = \hat{\sigma}^2 \left(\mat{A}+ \epsilon  \mat{I}_3\right)^{-1},
\end{equation}
where the value of $\hat{\sigma}^2$ can be estimated from the residuals on the moment in \eqref{eq:criterion1}:
\begin{equation}
	\label{eq:sigma_asip}
	{\hat{\sigma}}^2 = \frac{\sum_{i=1}^{N}  \left\| \vect{a}_i \times \p{}{asip}{} + \vect{b}^o_i \right\|^2}{N(3N-3)}.
\end{equation}
Similar to the AVOF, the main directions of variation can be extracted using an SVD on the covariance matrix $\mat{C}_{asip}$:
\begin{equation}
	\label{eq:svd_asip}
	\mat{C}_{asip} = \mat{U}_{asip} ~ \mat{\Sigma}_{asip} ~ \mat{U}_{asip}^T,
\end{equation}
with $\mat{U}_{asip}$ the matrix of singular vectors and $\mat{\Sigma}_{asip}$ a diagonal matrix containing the singular values $\mat{C}_{asip}$.
The first column of $\mat{U}_{asip}$ is interpreted as the direction in which the estimation of the ASIP is the most uncertain. Fig.~\ref{fig:asip} illustrates the ASIP for a set of screws where the corresponding directions of variation are illustrated by an uncertainty ellipsoid.

The calculation of the ASIP position, $\p{}{asip}{}$, the singular values $\mat{\Sigma}_{asip}$, the covariance matrix $\mat{C}_{asip}$, and the matrix of singular vectors $\mat{U}_{asip}$ will be referred to by the function operator $\texttt{asip}$ in the rest of this article. Algorithm \ref{alg:asip} summarizes the ASIP method.

\begin{algorithm2e}[t!]
	\footnotesize
	\caption{\color{black}Average Screw-axes Intersection Point\color{black}}
	\label{alg:asip}
	\SetKwFunction{ASIP}{asip}
	\SetKwProg{Fn}{Function}{:}{}
	\Fn{\ASIP{\color{black} $\{\screw{s}_i\}$\color{black}}}{
		\KwIn{ \color{black} $\{\screw{s}_i\}$ \color{black} with $i = 1 \dots N$ \tcp*{\color{black}set of screw axes\color{black}}\\
		}
		\KwOut{%
			$\vect{p}_{asip}$, $\mat{C}_{asip}$ \; \\
		}
		$\mat{A} \gets \frac{1}{N} \sum_{i=1}^{N}   [{\vect{a}}_i]_{\times} {[\vect{a}_i}]_{\times}^T  $ \;
		$\vect{p}_{asip} \gets \left( \mat{A} + \epsilon  \mat{I}_3 \right)^{-1} \left[  \frac{1}{N} \sum_{i=1}^{N}  ([{\vect{a}}_i]_{\times} \vect{b}^o_i)  + \epsilon \vect{p}_0 \right]$ \;
		${\hat{\sigma}}^2 \gets \frac{\sum_{i=1}^{N}  \left\| \vect{a}_i \times \p{}{asip}{} + \vect{b}^o_i \right\|^2}{N(3N-3)}$ \;
		$\mat{C}_{asip}$ $\gets$ ${\hat{\sigma}}^2 \left( \mat{A} + \epsilon  \mat{I}_3 \right)^{-1}$ \tcp*{covariance matrix} 
	}
	\normalsize
\end{algorithm2e}

\textit{Relation between AVOF and ASIP:}
When the AVOF method is applied to the directional component $\{\vect{a}_i\}$ of a set of screws $\{\screw{s}_i\}$, the following relationship can be established between the covariance matrix of the AVOF and the covariance matrix of the ASIP \cite{ancillao2022optimal}:
\begin{equation}
	\boldsymbol{C}_{asip} = \frac{\hat{\sigma}^2}{\operatorname{trace}(\mat{C}_c)} ( \mat{I}_3 - \boldsymbol{C}_{avof} )^{-1}.
\end{equation}
Consequently, the AVOF and ASIP have the same directions of variation since they share the same set of singular vectors. Additionally, from this equation it follows that the highest degree of uncertainty for the ASIP will always be in the dominant direction of the screw axes, as also visualized in Fig.~\ref{fig:asip}.

\section{Derivation of the Task Frame}
\label{sec:task_frame_derivation}

This section outlines the proposed method to derive the task frame. We first delineate the input data and expected outcome. Then the global approach is explained, followed by the detailed procedures to select the origin and orientation of the task frame. Finally we provide details on how to represent the demonstration data in the obtained task frame.

\subsection{Input Data and Expected Outcome}
\label{sec:input_data_and_expected_outcome}

The inputs for deriving the task frame are the motion and wrench data collected during one or more demonstration trials. To ensure a representative task frame, care has to be taken that the demonstrations sufficiently excite the degrees-of-freedom (d.o.f.) in motion and interaction wrench that are expected to be present in the actual performance of the learned task by the robot. Typically, the data is represented as a function of discrete time. We assume the data are segmented, i.e. they start at the beginning and finish at the end of the motion in contact to be modeled. Other assumptions are that the measured wrench corresponds to the interaction wrench, i.e. gravity is compensated and inertial forces are negligible, and the measured poses are sufficiently smooth such that smooth twists signals can be derived.

In general, the task frame $\refframe{tf}$ can be a freely moving frame. However, this paper limits itself to finding the best possible task frame that has: (1) a fixed origin w.r.t. either a reference frame $\refframe{w}$ attached to the world $\mathcal{W}$ or w.r.t. a reference frame $\refframe{tl}$ attached to the tool $\mathcal{TL}$, and (2) a fixed orientation w.r.t. either $\refframe{w}$ or $\refframe{tl}$. We refer to this as selecting the \textit{viewpoints}, \color{black} $\mathcal{W}$ or $\mathcal{TL}$, \color{black} for the origin and the orientation. Despite these limitations, such an approach is already suitable for a wide variety of practical applications, as abundantly shown in literature and demonstrated in Sections \ref{sec:experiments} and \ref{sec:exp_results}.

Consequently, the expected outcome of the task frame derivation procedure consists of: (1) the viewpoint for the origin ($\mathcal{W}$ or $\mathcal{TL}$) and the position vector representing the origin in the corresponding reference frame ($\refframe{w}$ or $\refframe{tl}$), i.e. $\p{tf}{*}{}$ where `$*$' is $w$ or $tl$, and similarly, (2) the viewpoint for the orientation and the rotation matrix (or other representation) representing the orientation w.r.t. the corresponding reference frame, i.e. $\R{tf}{\circ}{}$ where `$\circ$' is $w$ or $tl$.

With the task frame defined, the motion and wrench data can be represented in this frame to serve as a starting point for specifying desired values for the controller. 
To allow representation of the data in a time-independent way, a suitable geometric progress variable $\xi$ has to be chosen, as explained in Section \ref{sec:progress}. Typically, the progress rate $\dot{\xi}$ is chosen based on the instantaneous translational velocity, rotational velocity or a combination of both. This choice is made depending on the application and is based on an expert's insight. Therefore, as a third outcome of the procedure, we aim to determine an appropriate progress rate for the application in a data-driven way. 

The whole procedure should deliver invariant outcomes, i.e. results that are independent of the chosen data representation, and involve as few hyperparameters as possible to limit the required expert knowledge to tune them. If hyperparameters are needed, they are preferably as much as possible application-independent, to limit the tuning effort across different applications. Also, the outcomes should be easily transferable from the demonstration setup to the robot setup while adequately dealing with the differences between both.

\subsection{Conceptual Approach}
\label{sec:conceptual_approach}
\color{black} Inspired by the related work reviewed in Section \ref{subsec: selection_tf}, \color{black} we hypothesize that the underlying aim of an expert behind the choice of the task frame is to achieve decoupled control according to two principles: (1) decoupling between the directional and moment component of the motion and wrench screws, and (2) decoupling between motion and wrench.

\subsubsection{Principle 1: Decoupled Screw Direction and Moment} 
\label{sec:decoupled_screw_direction_and_moment}

Decoupled control between the \color{black} directional component $\vect{a}$ and moment component $\vect{b}^o$ of a screw, where the screw is either a twist or a wrench, \color{black} can be achieved by choosing the reference point for the moment component of the screw to be on the line of the screw axis itself. This reference point is then taken as the origin of the task frame. The advantage is that control corrections in the directional component of the screw performed at this point will not affect the moment component of the screw. 

Following Principle 1, we propose an approach to automatically derive the origin of the task frame with its viewpoint. This approach starts from two ideal models in which a perfect decoupling can be achieved and determines which ideal model best fits the motion and the wrench data. 

We start by defining the two ideal models for the motion: 

\begin{figure}[!t]
	\centering
	\subfloat[]{\includegraphics[height = 2.5cm]{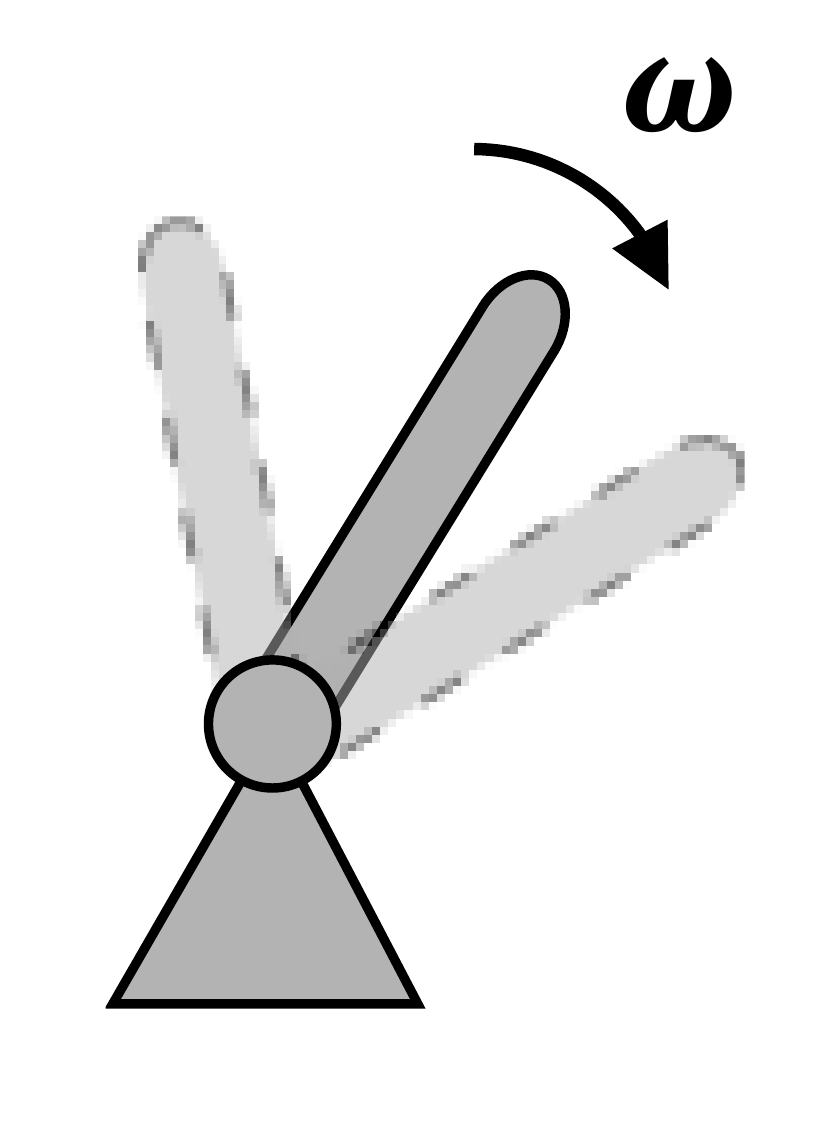}
		\label{fig:decouple_a}}
	\quad
	\subfloat[]{\includegraphics[height = 2.5cm]{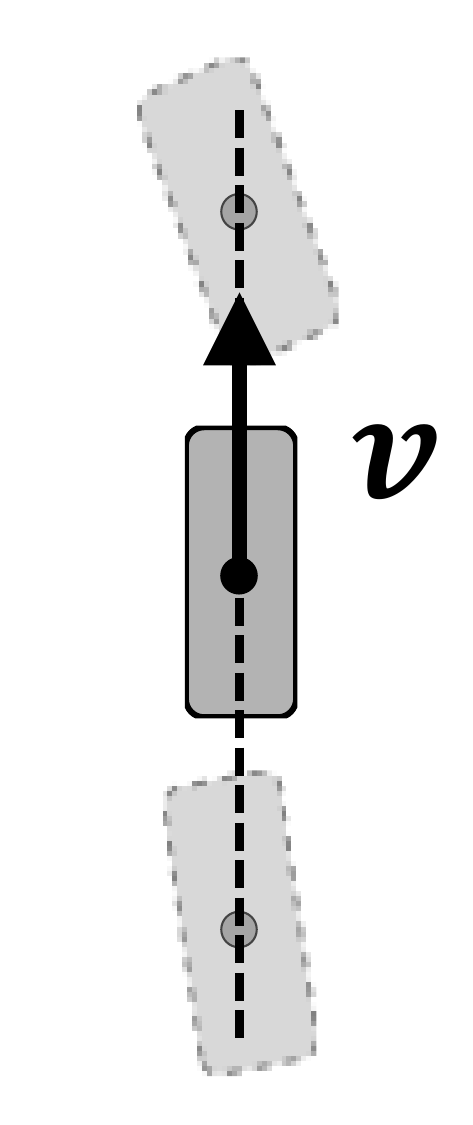}
		\label{fig:decouple_b}}
	\quad
	\subfloat[]{\includegraphics[height = 2.5cm]{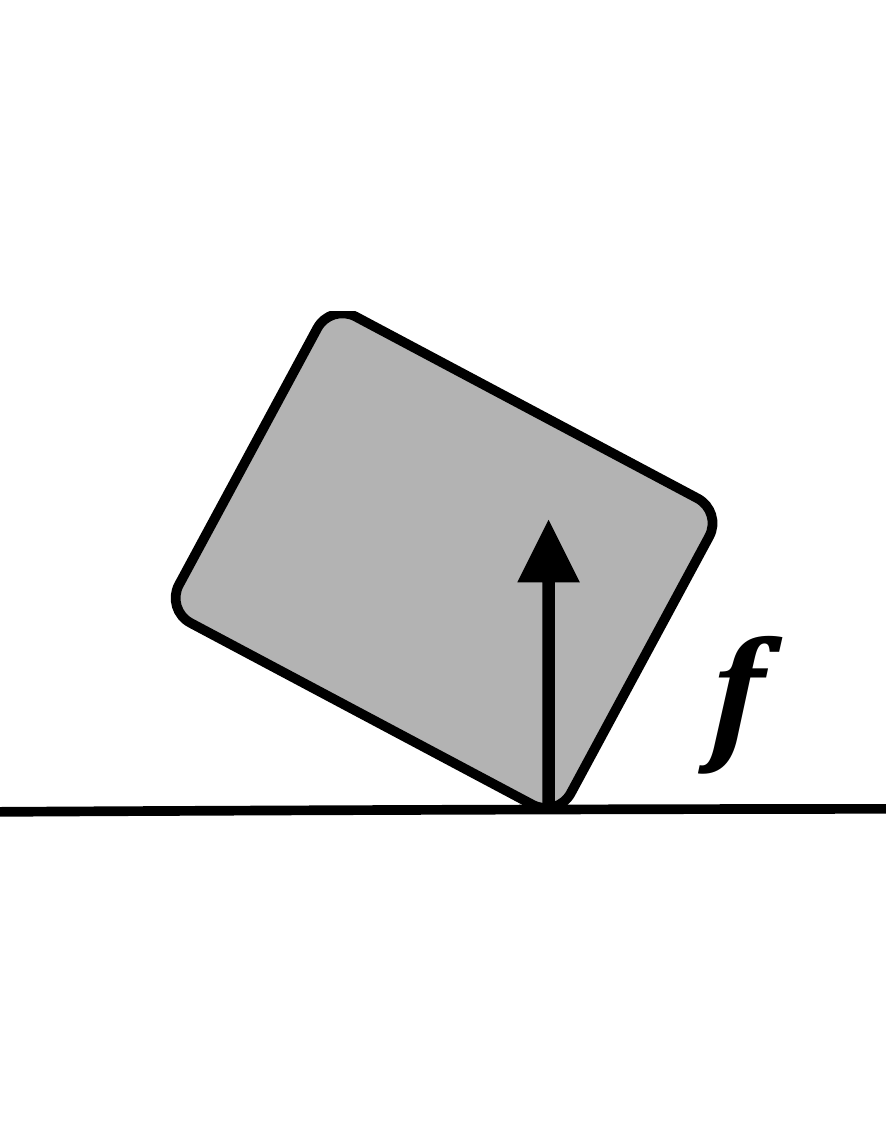}
		\label{fig:decouple_c}}
	\quad
	\subfloat[]{\includegraphics[height = 2.5cm]{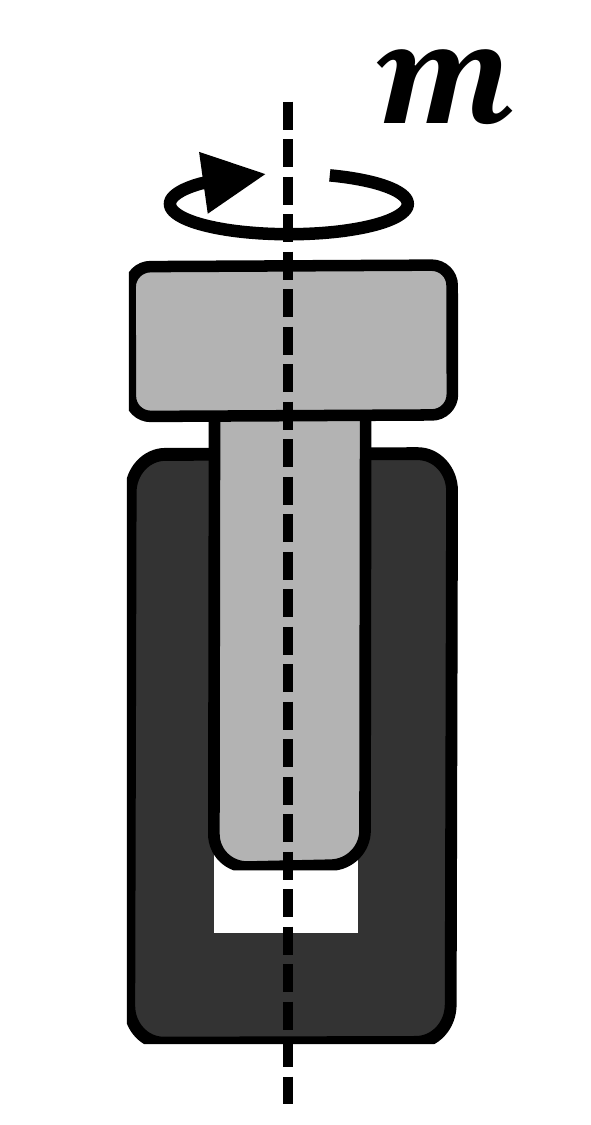}
		\label{fig:decouple_d}}
	\caption{Four models with a perfect decoupling between the direction and moment components of a screw: (a) pure rotation about fixed point; (b) constant translation of point; (c) pure force through fixed point; (d) constant moment through fixed point.
		\label{fig:decouple}}
\end{figure}


\textit{Model 1: pure rotation about fixed point, Fig.~\ref{fig:decouple_a}.} Consider a rigid body that is rotating about a point (e.g. a spherical joint) or about a line (e.g. a hinge). By choosing the origin in that
point or on that line, the translational velocity will be zero. Errors in the orientation can therefore be controlled without affecting the translation.

\textit{Model 2: constant translation of point, Fig.~\ref{fig:decouple_b}.} Consider a rigid body that performs a constant translation together with a general rotation about a reference point on the body. By choosing the origin in that point on the body, correcting an error in orientation about that point will not affect the translational motion of the point.

For the wrench, two similar models are defined:

\textit{Model 1: pure force through fixed point, Fig.~\ref{fig:decouple_c}.} Consider a contact between two rigid bodies where the wrench consists of a pure force through a point (e.g in a point-plane contact). By choosing the origin in that point the moment will be zero. Correcting an error in force through that point does not affect the moment about the point.

\textit{Model 2: constant moment through fixed point, Fig.~\ref{fig:decouple_d}.} Consider a contact between rigid bodies where the wrench consists of a constant moment together with a general force through a given point on the body. By choosing the origin in that point, correcting an error in force through that point does not affect the moment about the point.

We then apply the ASIP procedure introduced in Section \ref{sec:prelim_asip} to both the twist and the wrench screws to find out which of Models 1 and 2 best fits these data. For Model 1, the ASIP procedure is directly applied to screws $\screw{s}=\twist{}{}{}{}$ or $\wrench{}{}{}{}$. This results in the point with minimum average velocity or minimum average moment, respectively. In the ideal case where this minimum is zero, the motion and wrench data exactly correspond to their respective Model 1. In practice, this will never be the case, but the ASIP procedure results in the best possible origins to explain the motion and wrench data under the assumption of Model 1, together with their corresponding covariance matrices \eqref{eq:cov_asip}. 

For Model 2, the ASIP procedure is applied to the twist and wrench screws after subtracting their average: 
\begin{alignat}{1}
	\label{eq:def_screw_diff}
	\screwmv{s} = \screw{s} - \operatorname{mean}(\screw{s})
\end{alignat}
This eliminates the constant moment component of the screw; hence the remaining screw only has a directional component (with the average subtracted), and hence corresponds to the ideal case of Model 1. Consequently, this ASIP procedure results in the point with the most constant velocity or most constant moment, for motion and wrench respectively. In the ideal case where the deviation from a constant velocity or moment is zero, the motion and wrench data exactly correspond to Model 2. In practice this will never be the case, but the ASIP procedure results in the best possible origins to explain both motion and wrench data under the assumption of Model 2, together with their corresponding covariance matrices.

The criterion to decide which of Model 1 and 2 best fits the data is the comparison between the uncertainty of the two obtained origins, as given by the determinant of their covariance matrices, $\operatorname{det}(\mat{C}_{asip_1})$ and $\operatorname{det}(\mat{C}_{asip_2})$.

Section \ref{sec:task_frame_origin} provides more details about the derivation of the task frame origin.

As an additional result of this procedure, we can select a suitable progress rate to model the task: if Model 1  yields the best fit for the motion data, then the progress rate is best chosen as the rotational velocity, because the translational velocity of the origin of the task frame can become very small, hence is not suited; if Model 2 yields the best fit, then the progress rate is best chosen as the translational velocity of the origin, because this point has the most constant translational velocity.

\subsubsection{Principle 2: Decoupled Motion and Wrench} 
\label{sec:decoupled_motion_and_wrench}

Decoupled control of motion and wrench can be achieved by choosing the orientation for the task frame such that there is either motion or wrench control along and about the orthogonal axes of the frame. 

Historically, this approach received a lot of attention in the context of \textit{hybrid motion-force control}, which was conceived for \textit{ideal contacts}, i.e., frictionless contacts between rigid objects. However, many practical tasks in contact involve non-ideal contacts, and hence other controllers have been devised, such as admittance and impedance controllers, which combine motion and wrench control in the same direction. In our view, this lessens the need for separate control types in different directions. Still, even for these controllers a well-chosen orientation of the task frame can be helpful to specify possibly varying stiffness and damping matrices with different values in different directions.

Following Principle 2, we propose to derive the optimal orientation of the task frame with its viewpoint by analyzing the main direction and its variation of different 3-vector candidates $\vect{c}$ using the AVOF procedure introduced in Section \ref{sec:prelim_avof}. Possible inputs to the AVOF procedure are rotational velocity $\vect{\omega}$, force $\vect{f}$, translational velocity $\vect{v}$ and moment of force $\vect{m}$. The latter two have to be expressed at the origin resulting from Principle 1. From these four candidates we can select suitable \textit{vectors of interest} following the line of reasoning of Principle 1: if Model 1 yields the best fit to the motion or wrench data, then it is not very relevant to analyze the orientation of the translational velocity or moment of force and their variation, because they are very small at the origin, hence we prefer to analyze the rotational velocity or force. Conversely, if Model 2 yields the best fit, then it is more relevant to analyze the orientation of the (almost) constant translational velocity or moment and its variation, and derive the orientation of the task frame accordingly. This is summarized in Table \ref{tab:vectors_interest}.

Section \ref{sec:task_frame_orientation} provides more details about the derivation of the task frame orientation.

\begin{table}[!t]
	\centering
	\caption{Vectors of interest for determination of orientation and geometric progress variable}
	\label{tab:vectors_interest}
	\resizebox{\linewidth}{!}{
		\begin{tabular}{@{}lclcc@{}}
			\toprule
			\multirow{2}{*}{\textbf{Data}}
			& \multirow{2}{*}{\textbf{Criterion}}
			& \multirow{2}{*}{\textbf{Model}}
			& \multirow{2}{*}{\shortstack{ \textbf{Vectors of Interest} \\ (\textbf{Progress Variable} $\xi$) }} \\ \\ [2pt]
			\toprule
			\multirow{2}{*}{Twist}
			& $\operatorname{det}(\mat{C}_{asip_1}) < \operatorname{det}(\mat{C}_{asip_2})$
			& 1: min. $\vect{v}$
			& $\vect{\omega} \ (\dot{\xi}:= \norm{\vect{\omega}})$ \\
			& $\operatorname{det}(\mat{C}_{asip_1}) > \operatorname{det}(\mat{C}_{asip_2})$
			& 2: const. $\vect{v}$
			& $\vect{v} \ (\dot{\xi}:= \norm{\vect{v}})$ \\
			\multirow{2}{*}{Wrench}
			& $\operatorname{det}(\mat{C}_{asip_1}) < \operatorname{det}(\mat{C}_{asip_2})$
			& 1: min. $\vect{m}$
			& $\vect{f}$ \\
			& $\operatorname{det}(\mat{C}_{asip_1}) > \operatorname{det}(\mat{C}_{asip_2})$
			& 2: const. $\vect{m}$
			& $\vect{m}$ \\
			\bottomrule
		\end{tabular}
	}
\end{table}

\begin{figure}[!t]
	\centering
	\includegraphics[width = \linewidth]{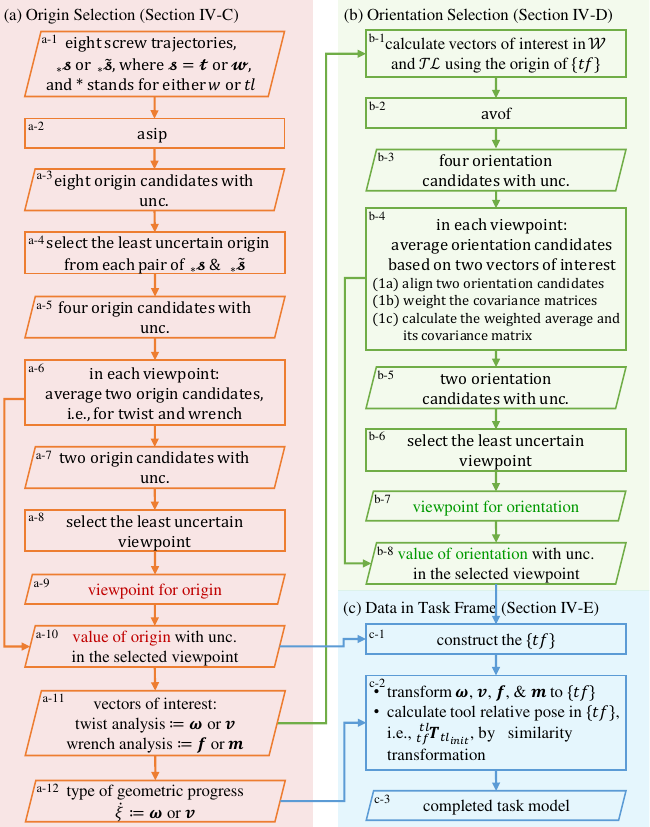}
	\caption{\color{black}Flow chart of the proposed approach. In this figure, unc. stands for uncertainty. (a) The origin of the task frame is derived, explained in section \ref{sec:task_frame_origin}. (b) The orientation of the task frame is derived, detailed in section \ref{sec:task_frame_orientation}. (c) The demonstration data including pose, twist, and wrench are expressed in the derived task frame as elaborated in section \ref{sec:task_frame_data}.\color{black}}
	\label{fig:alg_comp}
\end{figure}

\subsection{Selection of Origin \color{black}(Fig.~\ref{fig:alg_comp}a)\color{black}}
\label{sec:task_frame_origin}

As explained in Section \ref{sec:conceptual_approach} origin candidates for the task frame are found by applying the ASIP procedure to both twist and wrench data, and in two ways: to $\screw{s}$ and to $\screwmv{s}$.  In addition, to be able to decide about the viewpoint for the origin, we have to perform this procedure for screws that are expressed w.r.t. both the world frame $\refframe{w}$ and to the tool frame $\refframe{tl}$. Hence, in total eight applications of the ASIP procedure are required: to  $_{*}\screw{s}$ or $_{*}\screwmv{s}$, 
where $\screw{s}=\screw{t}$ or $\screw{w}$, and `$*$' stands for either $w$ or $tl$.
This yields eight origin candidates $\vect{p}_{asip}$ with their covariance matrix $\mat{C}_{asip}$. The eight candidates are reduced to a single origin in three steps. First, for each combination of a screw $\screw{s}$ and its mean-subtracted counterpart $\screwmv{s}$, we select the origin with the least uncertainty, i.e. with the smallest $\operatorname{det}(\mat{C}_{asip})$. Note that the unit of this uncertainty metric is always $[\text{m}^6]$ and that this step boils down to selecting between Model 1 or Model 2 for both twist and wrench screws in the two viewpoints. 
Next, since the uniform metric allows us to combine the results of twist and wrench analyses, the remaining four origin candidates are averaged two by two in each viewpoint, weighted by the inverse of their covariance matrix. I.e., given candidates $\vect{p}_{asip_t}$ and $\vect{p}_{asip_w}$ with corresponding $\mat{C}_{asip_t}$ and $\mat{C}_{asip_w}$, the averaged origin and its covariance matrix correspond to: 
\begin{align}
	\vect{p}_{avg}= \mat{C}_{avg} \left( \mat{C}_{asip_t}^{-1} \ \vect{p}_{asip_t} + \mat{C}_{asip_w}^{-1} 	 \ \vect{p}_{asip_w} \right),
	\label{eq:averaged point}
\end{align}
and:
\begin{align}
	\mat{C}_{avg}^{-1} = \mat{C}_{asip_t}^{-1} + \mat{C}_{asip_w}^{-1}.
	\label{eq:cov_orig}
\end{align}
Finally, to select the least uncertain viewpoint, $\operatorname{det}(\mat{C}_{avg})$ is again used to decide between the averaged origins in the two viewpoints.

So, the outcomes of this procedure are a viewpoint for the origin, a value for the origin and its covariance matrix.
Backtracking to the two selections made between the ASIP results in box a-4 for the selected viewpoint, we can further decide about the selection of the progress rate $\dot{\xi}$ and about the selection of the vectors of interest for the determination of a suitable orientation of the task frame, according to Table \ref{tab:vectors_interest}. 
No thresholds or other parameters are involved in the whole procedure.

\subsection{Selection of Orientation \color{black}(Fig.~\ref{fig:alg_comp}b)\color{black}} 
\label{sec:task_frame_orientation}

As explained in Section \ref{sec:conceptual_approach}  orientation candidates for the task frame are found by applying the AVOF procedure to the vectors of interest which result from the origin selection procedure in Section \ref{sec:task_frame_origin}: either $\vect{\omega}$ or $\vect{v}$ for twist, and either $\vect{f}$ or $\vect{m}$ for wrench. Values of $\vect{v}$ and $\vect{m}$ need to correspond to the translational velocity and moment of force at the selected task frame origin. This change to reference point $tf$ is achieved using \eqref{eq:refpoint_transf}. In addition, to be able to decide about the viewpoint for the orientation, we have to perform this procedure twice, once with all vectors expressed w.r.t. $\refframe{w}$ and once with all vectors expressed w.r.t. $\refframe{tl}$. Hence, four applications of the AVOF procedure are required: to  
$_{\circ}\vect{c}$, where $\vect{c}$ is either $\vect{\omega}$ or $\vect{v}$, and either $\vect{f}$ or $\vect{m}$, and `$\circ$' stands for $w$ or $tl$. 

The AVOF procedure yields four orientation candidates $\mat{U}_{avof}$ with their covariance matrix $\mat{C}_{avof}$ as found from \eqref{eq:svd_avof} and \eqref{eq:nondim1}, respectively. These four candidates are reduced to a single orientation in two steps: (1) averaging the two candidates resulting from either $\vect{\omega}$ or $\vect{v}$ and either $\vect{f}$ or $\vect{m}$ in every viewpoint and (2) selecting the viewpoint with the least resulting uncertainty. The first step is further divided into three substeps: (1a) aligning the two orientation candidates, (1b) weighting the covariance matrices, and (1c) calculating the weighted average and its covariance matrix. These substeps are summarized below.

\begin{figure}[!t]
	\centering
	\subfloat[]{\includegraphics[width = 2.45cm,page = 1]{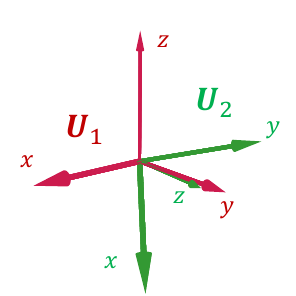}
		\label{fig:permutation_a}}
	\subfloat[]{\includegraphics[width = 2.45cm,page = 2]{figures/permutation.pdf}
		\label{fig:permutation_b}}
	\subfloat[]{\includegraphics[width = 2.45cm,page = 3]{figures/permutation.pdf}
		\label{fig:permutation_c}}
	\caption{Role of alignment and calculation of weighted average of two orientation frames $\mat{U}_1$ and $\mat{U}_2$. (a) The two orientation frames $\mat{U}_1$ and $\mat{U}_2$. (b) The two orientation frames after  alignment resulting in $\mat{R}_1$ and $\mat{R}_2$ . (c) The average of $\mat{R}_1$ and $\mat{R}_2$ is shown in black.}
	\label{fig:permutation}
\end{figure}

\textit{(1a) Aligning two orientation candidates.}
In every viewpoint, the two AVOF procedures result in two rotation matrices, $\mat{U}_1$ and $\mat{U}_2$, from twist and wrench data, respectively. The purpose is to find the best correspondence between the columns of $\mat{U}_1$ and $\mat{U}_2$ and label the corresponding pairs with $x$, $y$ or $z$ such that we obtain two rotation matrices $\mat{R}_1$ and $\mat{R}_2$ that are more or less aligned, as illustrated in Fig.~\ref{fig:permutation_a}, \ref{fig:permutation_b}.

Without loss of generality we choose the first orientation candidate as $\mat{R}_1 = \mat{U}_1$. Consequently, its $x$-axis is chosen according to the direction of the average $\vect{\omega}$ or $\vect{v}$, i.e. the main motion direction. Then the columns of $\mat{U}_2$ are reordered while assigning a proper sign to them to maintain righthandedness, such that the resulting orthonormal matrix corresponds to the rotation matrix $\mat{R}_2$ with the closest distance to $\mat{R}_1$. Algorithm \ref{alg:permutation} provides the detailed procedure. Note that this procedure also supports the case where the average direction of $\vect{f}$ or $\vect{m}$ is (almost) parallel to $\vect{v}$ or $\vect{\omega}$, i.e. also along the $x$-axis, as opposed to (almost) orthogonal as in the hybrid control paradigm. 
Also, this procedure does not affect $\mat{C}_{avof_2}$, because it remains defined by the corresponding input vectors $\vect{c}$, which are still expressed w.r.t. the same reference frame.

\begin{algorithm2e}[!t]
	\caption{Align two sets of three orthonormal vectors resulting in two orientation frames with minimum distance
	}
	\label{alg:permutation}
	\footnotesize
	\KwIn{$\mat{U}_1$, $\mat{U}_2$
	} 
	\KwOut{$\mat{R}_1$, $\mat{R}_2$
	}
	$\mat{R}_1 \gets \mat{U}_1$ \;
	$\mat{R}_\Delta \gets \mat{U}^{T}_{2} ~ \mat{R}_{1}$ \;
	\tcp{$c$ is a column index , $r$ is a row index }	
	\tcp{$\mat{P}$ permutates and adapts signs of columns. }
	\For{$c = 1 \dots 3$}{
		$r \gets \arg\max ~~ \operatorname{abs}\big(\mat{R}_\Delta(:,c)\big)$ \; 
		$\mat{P}(:,c) \gets 0$\;
		$\mat{P}(r,c) =  \operatorname{sign}(\mat{R}_\Delta(r,c)) $ \;
		\tcp{ensure that row cannot be selected in subsequent columns:}
		$\mat{R}_\Delta(r,:) \gets 0 $ \; 		
	}
	$\mat{R}_2 = \mat{U}_2 \mat{P}\ $;
	\normalsize	
\end{algorithm2e}

\textit{(1b) Weighting the nondimensional covariance matrices.}
Using the nondimensional $\mat{C}_{avof_1}$ and $\mat{C}_{avof_2}$ resulting from \eqref{eq:nondim1}  to calculate the average orientation of $\mat{R}_1$ and $\mat{R}_2$ has a benefit: it is completely data-driven, hence it does not involve any hyperparameters. However, the downside is that both rotation matrices are weighted according to how strong the vectors of interest $\vect{c}$ vary w.r.t. their average magnitude, which is not desired if this average magnitude is very small. An example is: very small, hence negligible contact forces measured in a human demonstration. In such case, the obtained orientation of these forces is irrelevant for the task model and should be discarded to avoid undesired bias in the averaged orientation. This can be achieved in an ad-hoc way based on an expert's insight, or automatically by applying an \textit{additional} weighting, i.e. after applying \eqref{eq:nondim1}. One approach is to choose the weights according to:
\begin{equation}
	\mat{C}_{avof} \gets \mat{C}_{avof}\frac{c_{ref}^2}{c_{avgsn}} 
\end{equation}
where $c_{ref}$ is a reference value for the magnitude of $\vect{c}$ below which, for example, the magnitude of the vector of interest is considered to be irrelevant, while $c_{avgsn}$ is the average of the squared norm of $\vect{c}$, which was also used in \eqref{eq:nondim1}. Such reference values are chosen for each of $\vect{\omega}$, $\vect{v}$, $\vect{f}$ and $\vect{m}$, which introduces four hyperparameters.

\textit{(1c) Calculating the weighted average.}
Given $\mat{R}_1$ and $\mat{R}_2$ with their resulting nondimensional and (optionally) weighted $\mat{C}_{avof_1}$ and $\mat{C}_{avof_2}$, the averaged orientation $\mat{R}_{avg}$ shown in black in Fig.~\ref{fig:permutation_c} can be found using the iterative procedure in Algorithm \ref{alg:wei_ave_orien}. 
This algorithm was first proposed for quaternions and dual quaternions in \cite{kavan2006dual} but was adapted here to rotation matrices. 
See Appendix~\ref{sec:appendix_combining} for the details.
The covariance of $\mat{R}_{avg}$ is calculated according to \eqref{eq:combining_covariance}. 

Finally, the viewpoint for the orientation is selected by comparing the determinant of the resulting covariance matrix and choosing the viewpoint with the smallest determinant.

So, the outcomes of this procedure are a viewpoint for the orientation, a value for the orientation and its covariance matrix. If there is no additional weighting in step (1b), no hyperparameters are involved in the whole procedure. Additional weighting introduces four hyperparameters, but they can be chosen intuitively and identical for a class of comparable applications.

\begin{algorithm2e}[!t]
	\caption{Weighted average of two rotations}
	\label{alg:wei_ave_orien}
	\footnotesize
	\KwIn{$\mat{R}_1$, $\mat{R}_2$, $\mat{C}_1$, $\mat{C}_2$
	} 
	\KwOut{$\mat{R}_{avg}$
	}
	$\mat{R} \gets \mat{R}_1 $ \tcp*{initialize}
	${\mat{\Lambda}}_1 \gets \left(\mat{C}_1^{-1} + \mat{C}_2^{-1}\right)^{-1} \mat{C}_1^{-1} $ \tcp*{weight 1}
	${\mat{\Lambda}}_2 \gets \left(\mat{C}_1^{-1} + \mat{C}_2^{-1}\right)^{-1} \mat{C}_2^{-1} $ \tcp*{weight 2}
	
	\Repeat{$\|{\vect{\delta}}\| < \epsilon \ $}{
		${\vect{\delta}} \gets {\mat{\Lambda}}_1 \log^{\vee} ( \mat{R}_1 {\mat{R}}^{-1} ) + {\mat{\Lambda}}_2 \log^{\vee} ( \mat{R}_2 {\mat{R}}^{-1})$\ \;
		$\mat{R} \gets \exp( {\vect{\delta}}) \mat{R}$ \tcp*{update rotation} 
	}	
	$\mat{R}_{avg} \gets \mat{R}$ \;
	\normalsize
\end{algorithm2e}

\subsection{Representation of Data in the Task Frame \color{black}(Fig.~\ref{fig:alg_comp}c)\color{black}}
\label{sec:task_frame_data}

From the origin $\p{tf}{*}{}$ and orientation $\R{tf}{\circ}{}$ of the task frame with their respective viewpoints, we can construct a transformation matrix for the task frame with respect to a general reference frame $\refframe{ref}$:
\begin{equation}
	\label{eq:transf_matrix_tf}
	\T{tf}{ref}{} =
	\begin{bmatrix}
		\R{\circ}{ref}{}\R{tf}{\circ}{} & \p{*}{ref}{} + \R{*}{ref}{}\p{tf}{*}{} \\
		\zerom{1\times3}  & 1
	\end{bmatrix}
\end{equation}
In the robot controller, typically $\T{tf}{w}{}$ or $\T{tf}{tl}{}$ are needed, i.e. with $ref=w$ or $tl$. Obviously, these matrices are only constant if the viewpoints of both origin and orientation correspond to $ref$. In other cases, the transformation matrix changes with the progress of the task. For example, for $ref=tl$, $*=tl$ and $\circ=w$, \eqref{eq:transf_matrix_tf} becomes:
\begin{equation}
	\T{tf}{tl}{} =
	\begin{bmatrix}
		\R{w}{tl}{}\R{tf}{w}{} &  + \p{tf}{tl}{} \\
		\zerom{1\times3}  & 1
	\end{bmatrix}
\end{equation}
where $\R{w}{tl}{}=\R{tl}{w}{}^{-1}$ is known from the instantaneous robot kinematics.

Then, we express all the demonstration data in $\refframe{tf}$. For twist $\screw{t}$ and wrench $\screw{w}$ this is achieved by transforming these screws to $_{tf}\screw{s}$ according to \eqref{eq:screw_transf}. For pose, the data is given by $\T{tl}{w}{}$ and can be decomposed into the initial pose of the tool, $\T{tl_{init}}{w}{}$, and the tool trajectory relative to that initial pose:
\begin{equation}
	\T{tl}{tl_{init}}{} =\T{tl_{init}}{w}{}^{-1} \T{tl}{w}{}
\end{equation}

By storing only the relative tool trajectory, the pose data becomes independent of the initial pose. This is very useful for transferring the skill from the demonstration environment to the robot, because usually the robot environment is different from the demonstration environment. For example, the world frame $\refframe{w}$ will typically not be at the camera or other sensor as during the demonstration, but instead at the robot base. Even if a model of this transformation is known, there might be calibration errors. Also, the object or workpiece that is fixed to $\refframe{w}$ could be placed at another location. All of this is taken care of by this approach. 
As an additional step, to be able to use the pose data in a controller that operates in the task frame, the pose trajectory $\T{tl}{tl_{init}}{}$ is expressed in the task frame using similarity transformation \eqref{eq:sim_transf}:
\begin{equation}
	\label{eq:sim_transf_applied}
	\T{tl}{tl_{init}}{tf} = \T{tf}{tl_{init}}{}^{-1} ~ \T{tl}{tl_{init}}{} ~ \T{tf}{tl_{init}}{}
\end{equation}
where generally $\T{tf}{tl_{init}}{}$ is a variable matrix given by
\begin{equation} 
	\T{tf}{tl_{init}}{}=\T{tl_{init}}{w}{}^{-1} \T{tf}{w}{}
\end{equation}
with $\T{tf}{w}{}$ obtained from \eqref{eq:transf_matrix_tf}.

So, the demonstration data is represented as $\twist{}{tf}{}{}$, $\wrench{}{tf}{}{}$ and $\T{tl}{tl_{init}}{tf}$. This completes the task frame derivation procedure. The demonstration data can be further processed, for example to align different demonstration trials by expressing them as a function of geometric progress $\xi$, and analyzed, for example by making statistical models based on the trials. This can be done in multiple ways, as in \cite{DMP2013},\cite{calinon2007learning}, and \cite{perico2019combining}, and is out of scope of this paper.

\section{Experiments Design}
\label{sec:experiments}

To validate the task frame derivation approach, it is subjected to a diverse range of applications. We anticipate that the approach will produce results closely, but not necessarily perfectly, aligned with expert choices. This expectation arises from the fact that in complex tasks, experts may rely on best guesses grounded in simplifying assumptions, and the algorithm's solution may surpass the expert's choice. We also want to investigate the effect of the optional weighting in the derivation of the task frame's orientation.

\color{black}To validate the benefits of using a well-chosen and parameterized task model, the task model obtained in Section \ref{sec:task_frame_data} \color{black}is utilized to perform \color{black}demonstrated tasks \color{black}by a robot. In pursuit of this, a general-purpose controller is designed based on the derived task frame. The controller's effectiveness is validated in the various applications. To assess its robustness and generalization capabilities, diverse scenarios involving disturbances or parameter variations are considered. To affirm the significance of utilizing a well-chosen task frame we investigate whether alterations in the task frame result in a deterioration of task performance. Additionally, controlling tasks within the derived task frame serves to validate that the demonstration data (twist, pose and wrench), represented as explained in Section \ref{sec:task_frame_data}, is accurate and sufficient, enabling accurate task performance by the robot.

\subsection{Applications}
\label{sec:exp_app}
\begin{figure*}[!t]
	\centering
	\subfloat[]{\includegraphics[width = 2.350cm, page = 1]{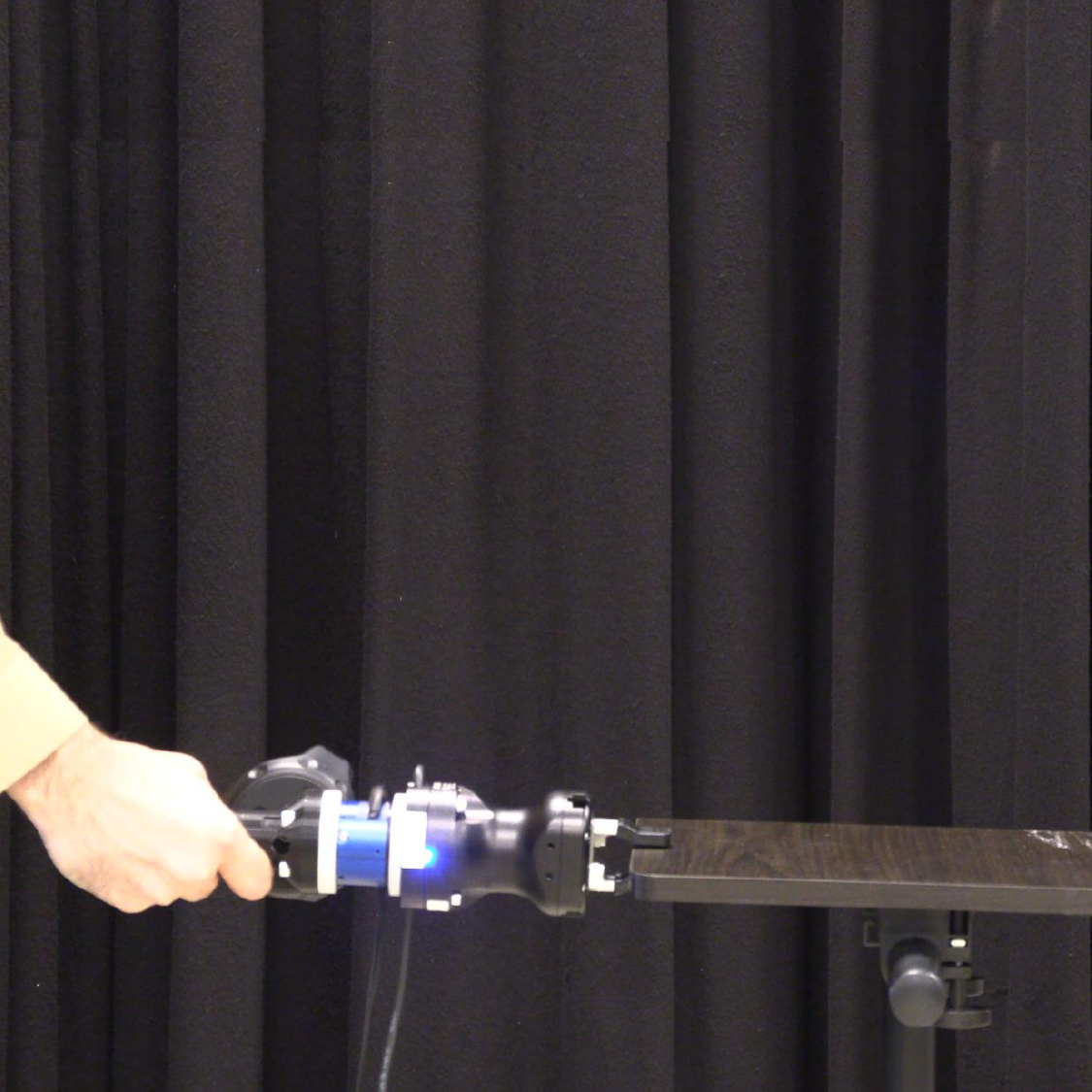}
		\label{fig:applications_zoom_a}}
	\hspace{0.02cm}
	\subfloat[]{\includegraphics[width = 2.350cm, page = 2]{figures/setups-zoom.pdf}
		\label{fig:applications_zoom_b}}
	\hspace{0.02cm}
	\subfloat[]{\includegraphics[width = 2.350cm, page = 3]{figures/setups-zoom.pdf}
		\label{fig:applications_zoom_c}}
	\hspace{0.02cm}
	\subfloat[]{\includegraphics[width = 2.350cm, page = 4]{figures/setups-zoom.pdf}
		\label{fig:applications_zoom_d}}
	\hspace{0.02cm}
	\subfloat[]{\includegraphics[width = 2.350cm, page = 5]{figures/setups-zoom.pdf}
		\label{fig:applications_zoom_e}}
	\hspace{0.02cm}
	\subfloat[]{\includegraphics[width = 2.350cm, page = 6]{figures/setups-zoom.pdf}
		\label{fig:applications_zoom_f}}
	\hspace{0.02cm}
	\subfloat[]{\includegraphics[width = 2.350cm, page = 7]{figures/setups-zoom.pdf}
		\label{fig:applications_zoom_g}}
	\caption{\color{black} Setups used for teaching different tasks to the robot: (a) Revolute Joint (b) Peg on Hole Alignment (c) Bottle Opening (d) Prismatic Joint (e) Drawing (f) 2-D Contour Following (g) 3-D Contour Following. \color{black} \color{black} The pictures are snapshots of the videos in the supplementary material. \color{black} The demonstration data for Bottle Opening and 2-D Contour Following are borrowed from \cite{perico2020learning}.}
	\label{fig:applications_zoom}
\end{figure*}

Fig.~\ref{fig:applications_zoom} depicts the seven tasks involving motion in contact considered in the experiments. 
Below we provide further explanation for each of them, paying particular attention to the task variability that was present in the demonstrations, which always consisted of five trials.

\subsubsection{Revolute Joint}
\label{sec:exp_revolute}

Interacting with a revolute joint such as when opening a door is a common daily activity.
This task has one d.o.f. for motion space and five d.o.f. for wrench space.
The tool is equipped with a gripper that can grip one body attached to the joint, in our case a table, and rotate it around the hinge. The setup is shown in Fig.~\ref{fig:applications_zoom_a}.
The operator demonstrated the task in different positions and with the joint in different horizontal orientations w.r.t. the world frame. The amount of rotation was almost the same in the different trials.

\subsubsection{Peg on Hole Alignment}
\label{sec:exp_peg}

This task can be used as an initialization before peg insertion in case it is difficult to provide a good estimate of the hole's location beforehand. First, we establish a three-point contact between the intentionally misaligned peg and the hole. Then, we rotate the peg until the symmetry axis of the peg is aligned with the symmetry axis of the hole. Unlike a simple revolute joint, this alignment involves a moving rotation axis as shown in \cite{bruyninckx1995peg}.
This task has three d.o.f. for both motion and wrench spaces. 
The peg is mounted on the tool and the hole is attached to the environment as shown in Fig.~\ref{fig:applications_zoom_b}.
The operator demonstrated the task in the same location and orientation w.r.t. the world frame. However, slight differences in the tool's orientation among demonstrations are observed due to human variability. The amount of initial misalignment is nearly the same among trials.

\subsubsection{Bottle Opening}
\label{sec:exp_bottle}

Removing the cap from a bottle is a non-reciprocal task, i.e., we need to apply moment and rotate the bottle opener in the same direction to remove the cap. Hence, unlike the previous tasks, this one requires non-negligible power consumption.
An opener is mounted on the tool and a bottle is fixed to the environment as shown in Fig \ref{fig:applications_zoom_c}. 
At the start of the task, initial contact is well-established, with the opener completely locked. Hence, there are zero and six d.o.f. in the motion and wrench spaces, respectively. During the opening phase, application of the interaction wrench causes motion, mainly but not exclusively in rotation.
The demonstration data for this application is borrowed from \cite{perico2020learning}.
The operator demonstrated the task in different positions, with slightly different orientations of the bottle's symmetry axis w.r.t. the vertical axis of the world frame. The amount of rotation is consistent among trials.

\subsubsection{Prismatic Joint}
\label{sec:exp_prismatic}

Interacting with a prismatic joint such as when opening a drawer is another common daily activity. 
This task has one d.o.f. for motion space and five d.o.f. for wrench space.
The same tool employed for interacting with a Revolute Joint was used, as shown in Fig.~\ref{fig:applications_zoom_d}.
The operator always demonstrated the task in the same position and orientation w.r.t. the world frame, with almost the same amount of translation.

\subsubsection{Drawing}
\label{sec:exp_drawing}

Tasks like drawing, writing and pizza cutting with a roller all involve a point-on-plane contact. The objective is to exert a force normal to a surface such as a table and simultaneously translate the tool over the table. In this study we considered drawing.
This task has one d.o.f. for wrench space and five d.o.f. for motion space. While the two translational d.o.f. along the table are functional during the demonstrations, the three orientations can be varied freely by the operator.
The tool is equipped with a pen-like rod as shown in Fig.~\ref{fig:applications_zoom_e}.
The operator demonstrated the task in the same location and orientation w.r.t. the world frame. There is not much variation in the tool's orientation w.r.t. the world frame and the plane of drawing remained the same among the trials.

\subsubsection{2-D Contour Following}
\label{sec:exp_2d}

The objective of this task, shown in Fig.~\ref{fig:applications_zoom_f}, is to follow a 2-D contour using a roller at almost constant translational velocity while maintaining an almost constant contact force. Polishing is a typical application.
The task has only one d.o.f. for wrench space, while there are five d.o.f. for motion space. One translational d.o.f. along the contour allows us to follow it. Since it is a 2-d.o.f. application, the tool is expected not to move in the two out-of-plane rotational directions and the out-of-plane translational direction, hence maintaining a constant orientation or position in those directions. Finally, the intention was to maintain also a constant in-plane orientation of the tool w.r.t. the world.
The tool consists of a rod with a roller at its tip. The demonstration data was again borrowed from \cite{perico2020learning}.
The operator demonstrated the task in different locations but with approximately the same orientation of the workpiece w.r.t. the world frame. Hence, also the plane of the contour remained the same among the trials.

\subsubsection{3-D Contour Following}
\label{sec:exp_3d}

The objective is to follow a 3-D contour with almost constant translational velocity while simultaneously exerting an almost constant amount of force normal to the contour's edge. 
The tool is shown in Fig.~\ref{fig:applications_zoom_g}, see \cite{vochten2023invariant} for more details. It has five wheels which are expected to be in contact during the experiment. The loss of contact for any of these wheels is considered a task failure. 
The task involves one instantaneous d.o.f. in motion space, i.e. translation along the contour's edge, and five instantaneous d.o.f. in wrench space, i.e. normal forces on two sides of the contour and three moments. 
Compared to the previous tasks, this task required much more skill from the operator to complete successful demonstrations, as it was not evident to maintain contact at all wheels during the entire trials.
The operator demonstrated the task with the workpiece in the same position and orientation w.r.t. the world frame.

\subsection{Demonstration Setup}
\label{sec:exp_setup}
The HTC VIVE system was used for recording pose data in all applications. This system includes a tracker that was mounted on the tool and a camera for recording the tracker's pose. Its accuracy is in the order of a few millimeters and a few degrees. In contrast, different force/torque sensors were used in the various applications, as detailed in Table \ref{tab:FT_sensors} along with their data sheet specifications. The first three sensors were used for recording demonstration data (respectively for applications 1/2/4/5, 3/6 and 7), while the last one was employed for all tasks performed by the robot. All the sensors used in the demonstration setups record data at a frequency of 200~$\text{Hz}$. Furthermore, in applications 1 and 4, we used the Robotiq Hand-E gripper; it has two states: open and close. 

\begin{table}[!t]
	\centering
	\caption{Specifications of the force/torque sensors}
	\label{tab:FT_sensors}
	\resizebox{\linewidth}{!}{
		\begin{tabular}{@{}lccccccc@{}}
			\toprule
			\multirow{3}{*}{\shortstack{ \textbf{6-Axis} \\ \textbf{Force/Torque sensor}}}
			& \multirow{3}{*}{\shortstack{\textbf{Nominal} \\ \textbf{Accuracy}}}
			& \multicolumn{6}{c}{\textbf{Measurement Range}} \\
			&
			& \multicolumn{3}{c}{\textbf{Force} $[\text{N}]$}
			& \multicolumn{3}{c}{\textbf{Torque} $[\text{Nm}]$} \\
			&
			& \multicolumn{2}{c}{$f_x, f_y$}
			& $f_z$
			& \multicolumn{3}{c}{$\tau_{x}, \tau_{y}, \tau_{z}$}
			\\ [2pt]
			\toprule
			JR3 50M31A3
			& $1 \%$
			& \multicolumn{2}{c}{$\pm 100$}
			& $\pm 200$
			& \multicolumn{3}{c}{\textbf{$\pm 5$}} \\	
			JR3 67M25A3
			& $1 \%$
			& \multicolumn{2}{c}{$\pm 200$}
			& $\pm 400$
			& \multicolumn{3}{c}{\textbf{$\pm 12$}} \\	
			JR3 67M25T3
			& $1 \%$
			& \multicolumn{2}{c}{$\pm 400$}
			& $\pm 800$
			& \multicolumn{3}{c}{\textbf{$\pm 24$}} \\	
			Schunk FTN-AXIA80
			& $<2 \%$
			& \multicolumn{2}{c}{$\pm 150$}
			& $\pm 470$
			& \multicolumn{3}{c}{$\pm 8$} \\
			\bottomrule
		\end{tabular}
	}
\end{table}

\subsection{Pre-Processing of Demonstration Data}
\label{sec:pre_data}

The motion tracker records the pose of a frame $\refframe{tr}$ attached to the tracker w.r.t. the camera frame $\refframe{c}$, which corresponds to the world frame $\refframe{w}$. Simultaneously, a force/torque sensor records wrench measurements in the force/torque sensor frame $\refframe{ft}$. Both tracker and force/torque sensor are attached to the tool frame $\refframe{tl}$. Their relative poses were assumed to be known.

To start, we segmented the data to identify the motion-in-contact part of each demonstration trial. We opted for a straightforward approach based on signal magnitude, since segmentation was not the focus of this study. A first segmentation was performed based on the magnitude of the force $\vect{f}$ or moment  $\vect{m}$ to make sure there was contact. This implies prior weight compensation of the measured wrench. We additionally processed the wrench signals using a moving average smoother weighted with a Gaussian distribution.  Subsequently, we further segmented the data based on the magnitude of the rotational velocity $\vect{\omega}$ or translational velocity $\vect{v}$ to eliminate the parts without significant motion.

The pose data $\T{tl}{w}{}$ used in the task frame derivation procedure was then calculated from the measured tracker poses $\T{tr}{w}{}$ and the relative pose $\T{tr}{tl}{}$, while the twist data $_w\screw{t}$ followed from 
pose differentiation according to \eqref{eq:pose_differentiation}. Next, we augmented the five demonstration trials for each application to obtain a batch. Lastly, we calculated screws $\screw{s}$ and $\screwmv{s}$, cf.~\eqref{eq:def_screw_diff}, in both viewpoints as required for origin selection in Fig.~\ref{fig:alg_comp}.

\subsection{Task Frame Derivation and Data Post-Processing}
\label{sec:post_data}

Task frame derivation was applied according to Section \ref{sec:task_frame_derivation} for all applications. 
No regularization was applied in the ASIP procedure, i.e. $\epsilon$ was set to zero in \eqref{eq:criterion1_sol}.
Furthermore, following reference values were selected for the optional weighting of the orientation: $\omega_{ref} = 0.05~\sfrac{\text{rad}}{\text{s}}$, $v_{ref} = 0.005~\sfrac{\text{m}}{\text{s}}$, $f_{ref} = 1~\text{N}$, and $m_{ref} = 0.1~\text{Nm}$. 
No further tuning of these values was applied.

After transforming the data to the obtained task frame according to Section \ref{sec:task_frame_data}, the following signals are determined: $x,y,z$-positions and unit quaternions for pose, and the components of the twist and wrench screws.
The signals were made time-invariant and aligned \textcolor{black}{to handle differences in velocity profile} by reparameterizing them according to \eqref{eq:reparam4} as functions of the nondimensional geometric progress variable $\bar{\xi}$ that resulted from the task frame derivation, while using the same number of samples $N=100$ for all five trials.
Next, we fitted a cubic spline to these aligned signals to find a smooth average over the trials. The quaternions are renormalized after smoothing.

In a final step, $\bar{\xi}$ was replaced by $\bar{\xi}.\xi_{\textit{max,avg}}$ where $\xi_{\textit{max,avg}}$ is the average traveled distance or angle of the trials.
The resulting signals were saved as the desired reference signals expressed in the task frame.
\subsection{Task Performance Setup and Control Strategy}
\label{sec:robot_setup}

A UR10e robot is used to perform the tasks. The robot is equipped with a Schunk force/torque sensor, see Table \ref{tab:FT_sensors} for its specifications, and is controlled with a frequency of 100 $\text{Hz}$. 
We employed a constraint-based and velocity-resolved control framework known as \textit{eTaSL} \cite{aertbelien2014etasl}, and devised a single type of controller for all applications. Evidently, other types of controllers can be applied as well. In this approach desired values for pose, twist and wrench are combined in two sets of constraints that are applied simultaneously, as detailed below. 

Pose constraints are responsible for tracking the desired pose trajectory expressed in the task frame, $\T{tl}{tl_{init}}{tf}^{~des}$. 
The actual pose trajectory is also expressed in the task frame through a similarity transformation applied to the relative trajectory w.r.t. the initial pose, $\T{tl}{tl_{init}}{}^{act}$, in the same way as the desired trajectory in \eqref{eq:sim_transf_applied}, but now in real-time:
\begin{equation}
	\label{eq:sim_transf_act}
	\T{tl}{tl_{init}}{tf}^{act} = {\big(\T{tf}{tl_{init}}{}^{act}\big)}^{-1} ~ \T{tl}{tl_{init}}{}^{act} ~ \T{tf}{tl_{init}}{}^{act}
\end{equation}
The pose constraint ensures that the pose difference:
\begin{equation}
	\Delta \mat{T} = {\big( \T{tl}{tl_{init}}{tf}^{act} \big)}^{-1} \T{tl}{tl_{init}}{tf}^{des}
\end{equation}
evolves to the identity matrix $\mat{I}_4$ using a desired control twist:
\begin{equation}
	\label{eq:constraint_pose}
	\prescript{}{tf}{\screw{t}}_{contr,p} = ~ k_{p} \operatorname{log}^{\vee} \big( \Delta \mat{T} \big) + \prescript{}{tf}{\screw{t}}_{des}
\end{equation}
Here, the first term is the feedback term, comprising feedback gain $k_{p}$ multiplied by the displacement screw derived from the pose error $\Delta \mat{T}$ according to \eqref{eq:displacement_screw}.
The second term is a feedforward term which aims at reducing the reference tracking error.

Wrench constraints ensure a safe interaction between the tool and the environment and aim at tracking the desired wrench trajectories. This is expressed as a constraint, $_{tf}\screw{w}_{act} =  {}_{tf}\screw{w}_{des}$, which is also converted to a desired control twist: 
\begin{equation}
	\label{eq:constraint_wrench}
	\prescript{}{tf}{\screw{t}}_{contr,w} = ~ k_{w} \hat{\mat{C}}_w (\prescript{}{tf}{\screw{w}}_{des} - \prescript{}{tf}{\screw{w}}_{act}) + \prescript{}{tf}{\screw{t}}_{des}
\end{equation}
where $k_{w}$ is a feedback gain and $\underline{\Delta}\,\hat{\mat{C}}_w$ is a diagonal compliance matrix, where $\underline{\Delta} = \scriptsize
\begin{bmatrix} \mat{0}_3 &\mat{I}_3 \\
	\mat{I}_3& \mat{0}_3
\end{bmatrix}$.
They are multiplied by the wrench error to form the feedback term. The second term in \eqref{eq:constraint_wrench} is the same feedforward term as in \eqref{eq:constraint_pose}.

\begin{figure}[!t]
	\centering
	\includegraphics[width = 7.5cm]{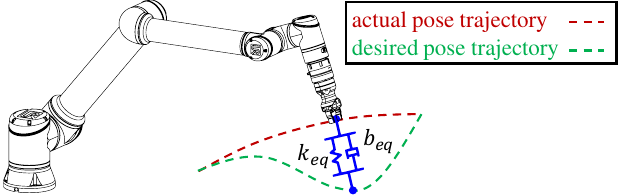}
	\caption{Schematic representation of the proposed constraint-based control strategy. Deviation from the desired pose trajectory occurs as a result of the wrench constraint. This can be modeled as a spring-damper system.}
	\label{fig:control}
\end{figure}

The feedforward terms in \eqref{eq:constraint_pose} and \eqref{eq:constraint_wrench} generate the exact required motion if the demonstration and robot scenarios are exactly the same. Hence, little or no contribution from the feedback terms is required in that case. If however the robot scenario deviates from the demonstration scenario, the feedback terms in the pose and wrench constraints come into play to compensate for these deviations. As a result, the desired control twists \eqref{eq:constraint_pose} and \eqref{eq:constraint_wrench} are no longer consistent, but conflicting.
To solve this conflict, weights $w_p$ and $w_w$ are applied to these equations, respectively, which results in a spring-damper like behavior, as depicted in Fig.~\ref{fig:control}. The equivalent stiffness and damping parameters, $k_{eq}$ and $b_{eq}$, are determined by gains $k_p$ and $k_w$, weights $w_p$ and $w_w$, and compliance $\underline{\Delta}\,\hat{\mat{C}}_w$.
\color{black}
This relation can be shown for a 1-d.o.f. translational system,
\begin{align}
	k_{eq} = \frac{w_{p} k_{p}}{w_{w} k_{w} \hat{C}_f} ~\text{and} ~~ b_{eq} = \frac{1}{w_{w} k_{w} \hat{C}_f}
\end{align}
where $\hat{C}_f$ is the translational compliance. \color{black}

Given the large number of setups, little effort was spent in parameter tuning. The control parameters used in every application are listed in Table \ref{tab:gains}. 
The same control gains $k_p = k_w = 3~\text{s}^{-1}$ were selected for all applications. These gains correspond to the achievable bandwidth expected from closed-loop control with robots like the UR10e, with a safety margin. We roughly estimated the diagonal entries of the compliance matrix $\underline{\Delta}\,\hat{\mat{C}}_w$ for every application set-up, as a simplified model for the true physical compliance between the tool and the environment, thereby ensuring stable closed-loop wrench control. We took the same value $\hat{C}_f$ for the three translational directions and the same value $\hat{C}_m$ for all three rotational directions. Furthermore, for each application we selected from three possible weights $w_p=0.01,~0.03\ \text{or} \ 0.1$, while $w_w=1-w_p$. So, only three parameters were selected for each application: rough estimates for $\hat{C}_f$  and $\hat{C}_m$, and one out of three options for $w_p$.

The performance of the controller proposed above is anticipated to be independent of the task frame orientation, because (1) the postprocessing of the demonstration data was limited to reparameterization and spline fitting, so no orientation-specific data was thrown away, and (2) all controller gains, compliance values and weights are equal in the three translational directions and in the three rotational directions. This will not be the case with more sophisticated controllers which aim at leveraging non-isotropic design. 
The decoupling introduced in principle 2 of section~\ref{sec:conceptual_approach} can help in designing such controllers. Such controllers are however out of scope of this paper.

\begin{table}[!t]
	\centering
	\caption{Control parameters used in the experiments.}
	\label{tab:gains}
	\resizebox{\linewidth}{!}{
		\begin{tabular}{@{}lcccccc@{}}
			\toprule
			\multicolumn{1}{c}{\multirow{2}{*}{\textbf{Task}}}
			& $k_p$
			& $k_w$
			& $\hat{C}_f$
			& $\hat{C}_m$
			& $w_p$
			& $w_w$ \\
			
			& $[\text{s}^{-1}]$
			& $[\text{s}^{-1}]$
			& $\text{e}^{-3}[\sfrac{\text{m}}{\text{N}}]$
			& $\text{e}^{-3}[\sfrac{\text{rad}}{\text{Nm}}]$
			& $[-]$
			& $[-]$	\\
			\toprule
			Revolute Joint
			& $3$
			& $3$
			& $0.4$
			& $0.7$
			& $0.01$
			& $0.99$ \\
			Peg on Hole Alignment
			& $3$
			& $3$
			& $0.7$
			& $0.7$
			& $0.10$
			& $0.90$ \\
			Bottle Opening
			& $3$
			& $3$
			& $0.1$
			& $100$
			& $0.01$
			& $0.99$ \\
			Prismatic Joint
			& $3$
			& $3$
			& $0.5$
			& $0.5$
			& $0.03$
			& $0.97$ \\
			Drawing
			& $3$
			& $3$
			& $0.5$
			& $200$
			& $0.10$
			& $0.90$ \\
			2-D Contour Following
			& $3$
			& $3$
			& $0.8$
			& $200$
			& $0.03$
			& $0.97$ \\
			3-D Contour Following
			& $3$
			& $3$
			& $0.1$  
			& $50$
			& $0.01$
			& $0.99$ \\
			\bottomrule
		\end{tabular}
	}
\end{table}

\subsection{Scenarios}
\label{sec:scenarios}

The proposed control approach is challenged in three types of scenarios, as outlined below.

\textit{Nominal scenarios.}
Here, the robot first performs the task under the same conditions as observed during the demonstrations, i.e. the tool's pose, the contact wrench and the task speed closely resemble what was extracted from the demonstrations. Subsequently, following features were investigated: role of the pose and wrench constraints, independence of the initial pose, and variation of task speed and contact wrench.

\textit{Changed geometry.}
If the tool is changed from the one used in the demonstrations, this may affect task frame parameters $\p{tf}{tl}{}$  and/or $\R{tf}{tl}{}$.  
Similarly, the geometry of the environment instead of the tool can be different from the one during the demonstration.
Two approaches were investigated: (1) changing the task frame parameters manually, based on physical insight, and (2) performing the task with the original parameters, recording the data and reintroducing them in the task frame derivation procedure to find improved parameters.

\textit{Sensitivity to task frame origin and orientation.}
We performed two experiments in which the task frame was changed manually w.r.t. the optimal one that resulted from the procedure proposed in Section \ref{sec:task_frame_derivation}: a change of orientation and a change of origin.

\section{Experimental Results}
\label{sec:exp_results}

Video recordings of all demonstrations and all robot task performances can be found in the supplementary material.\footnote{The Matlab codes and the demonstration data will be available as a Git repository with the final version of the paper.}

\subsection{Task Frame Derivation}
\label{sec:exp_task_frame}
Table \ref{tab:results_approach} summarizes the results of the task frame derivation. Columns 2-5 contain the results of the discrete decisions: viewpoints for the origin and the orientation \color{black} ($\mathcal{W}$ or $\mathcal{TL}$), \color{black} and the vectors of interest for motion and wrench \color{black} ($\vect{v}$ or $\vect{\omega}$, and $\vect{f}$ or $\vect{m}$)\color{black}. Each decision is based on the comparison of two relevant covariance matrices $\mat{C}_1$ and $\mat{C}_2$. Therefore, \color{black} between brackets, \color{black} we also report a ratio,
\begin{equation}
	\text{Ratio} =\sqrt{
		\frac{\operatorname{max}(\operatorname{det}(\mat{C}_{1}),\operatorname{det}(\mat{C}_{2}))}
		{\operatorname{min}(\operatorname{det}(\mat{C}_{1}),\operatorname{det}(\mat{C}_{2}))}
	},
	\label{eq:ratio}
\end{equation}
as a measure of the statistical significance of the decision. While a large value of the ratio corresponds to a large significance, a value close to 1 corresponds to a low significance. I.e., in the latter case the other option might work equally well in practice.

Columns 6-13 focus on the values derived for the origin and orientation of the task frame, $\refframe{tf}$, in comparison with the expert's choice for the task frame, $\refframe{tf^{*}}$. The expert choice was based on the authors' best judgment. Derived and expert task frames are visualized for each application in the video recordings in the supplementary material. The differences between the origins and the orientations of $\refframe{tf}$ and $\refframe{tf^{*}}$ were calculated in different ways, to capture their most relevant component(s) for each particular application. This is detailed at the bottom of the table. For example, for all applications except Drawing and 2-D Contour Following, the angle between the main axes of motion, $\e{[tf]}{}{x}$ and $\e{[tf^{*}]}{}{x}$, was chosen for evaluating the difference in orientation, while for the other two the angle between the normals to the plane of motion, $\e{[tf]}{}{z}$ and $\e{[tf^{*}]}{}{z}$, was chosen. The differences in origin and orientation are reported for four task frame candidates: the task frames resulting from the motion data, from the wrench data, their unweighted average and their weighted average. For each application, the smallest differences are marked in bold.

\begin{table*}[!t]
	\centering
	\caption{Outcome of the proposed approach for selection of the task frame and comparison with task frames chosen by an expert.\\ \color{black} The comparison with the expert for the Drawing task is visualized in Fig.~\ref{fig:task_frames}. For the rest of the tasks, please refer to the videos in the supplementary material for the visualization.\color{black} }
	\label{tab:results_approach}
	\resizebox{\linewidth}{!}{
		\begin{tabular}{@{}lcccccccccccc@{}}
			\toprule
			\multicolumn{1}{c}{ \multirow{3}{*}{\textbf{Task}} }
			& \multicolumn{2}{c}{ \textbf{Viewpoint} }
			& \multicolumn{2}{c}{ \textbf{Vectors of Interest} }
			& \multicolumn{8}{c}{ \textbf{Derived Task Frame $\refframe{tf}$ vs. Expert's Task Frame $\refframe{tf^{*}}$} } \\ 
			& \textbf{Origin}
			& \textbf{Orientation}
			& \textbf{Motion}
			& \textbf{Wrench}
			& \multicolumn{2}{c}{ \textbf{From Motion} }
			& \multicolumn{2}{c}{ \textbf{From Wrench} }
			& \multicolumn{2}{c}{ \textbf{Unweighted Avg.} }
			& \multicolumn{2}{c}{ \textbf{Weighted Avg.} } \\
			& \textbf{(Ratio)}
			& \textbf{(Ratio)}
			& \textbf{(Ratio)}
			& \textbf{(Ratio)}
			& $[^{\circ}]$
			& $[\text{mm}]$
			& $[^{\circ}]$
			& $[\text{mm}]$
			& $[^{\circ}]$
			& $[\text{mm}]$
			& $[^{\circ}]$
			& $[\text{mm}]$ \\ [2pt]
			\toprule
			Revolute Joint
			& $\mathcal{TL} \ (1.3\text{e}^{1})$ 
			& $\mathcal{TL} \ (1.9\text{e}^{1})$ 
			& $\vect{\omega} \ (6.9)$
			& $\vect{f} \ (2.3)$
			& $2.3^{\text{a}}$
			& $4.4^{\text{b}}$
			& $9.9^{\text{a}}$
			& $100.8^{\text{b}}$
			& $\mathbf{2.3}^{\text{a}}$
			& $\mathbf{4.4}^{\text{b}}$
			& $\mathbf{2.3}^{\text{a}}$
			& $4.5^{\text{b}}$ \\ 
			Peg on Hole Alignment
			& $\mathcal{TL} \ (1.9)$
			& $\mathcal{TL} \ (1.2)$
			& $\vect{\omega} \ (1.7\text{e}^{1})$ 
			& $\vect{f} \ (3.5)$
			& $2.6^{\text{a}}$
			& $15.9^{\text{b}}$
			& $4.8^{\text{a}}$
			& $42.0^{\text{b}}$
			& $\mathbf{2.0}^{\text{a}}$
			& $\mathbf{26.6}^{\text{b}}$
			& $2.6^{\text{a}}$
			& $27.5^{\text{b}}$ \\ 
			Bottle Opening
			& $\mathcal{TL} \ (2.8\text{e}^{1})$ 
			& $\mathcal{TL} \ (4.2)$
			& $\vect{\omega} \ (2.2)$
			& $\vect{m} \ (4.3)$
			& $2.7^{\text{a}}$
			& $18.5^{\text{b}}$
			& $12.1^{\text{a}}$
			& $180.8^{\text{b}}$
			& $\mathbf{10.8}^{\text{a}}$
			& $\mathbf{26.0}^{\text{b}}$
			& $12^{\text{a}}$
			& $26.2^{\text{b}}$ \\ 
			Prismatic Joint
			& $\mathcal{TL} \ (1.2)$
			& $\mathcal{TL} \ (1.3)$
			& $\vect{v} \ (1.3\text{e}^{1})$ 
			& $\vect{f} \ (1.6)$
			& $1.3^{\text{a}}$
			& $358.7^{\text{b}}$
			& $37.1^{\text{a}}$
			& $122.9^{\text{b}}$
			& $\mathbf{1.3}^{\text{a}}$
			& $\mathbf{166.5}^{\text{b}}$
			& $1.3^{\text{a}}$
			& $167.6^{\text{b}}$ \\ 
			Drawing
			& $\mathcal{TL} \ (1.7\text{e}^{3})$ 
			& $\mathcal{W} \ (9.5)$
			& $\vect{v} \ (2.0)$
			& $\vect{f} \ (2.6)$
			& $0.3^{\text{c}}$
			& $498.8^{\text{d}}$
			& $3.9^{\text{c}}$
			& $8.9^{\text{d}}$
			& $3.8^{\text{c}}$
			& $\mathbf{8.9}^{\text{d}}$
			& $\mathbf{3.7}^{\text{c}}$
			& $\mathbf{8.9}^{\text{d}}$ \\ 
			2-D Contour Following
			& $\mathcal{TL} \ (6.8\text{e}^{2})$ 
			& $\mathcal{W} \ (1.1)$
			& $\vect{v} \ (1.6)$
			& $\vect{f} \ (1.9)$
			& $4.1^{\text{c}}$
			& $344.1^{\text{e}}$
			& $5.9^{\text{c}}$
			& $2.0^{\text{e}}$
			& $8.5^{\text{c}}$
			& $\mathbf{2.0}^{\text{e}}$
			& $\mathbf{6.2}^{\text{c}}$
			& $\mathbf{2.0}^{\text{e}}$ \\ 
			3-D Contour Following
			& $\mathcal{TL} \ (1.1\text{e}^{3})$ 
			& $\mathcal{TL} \ (1.8\text{e}^{1})$ 
			& $\vect{v} \ (6.1\text{e}^1)$ 
			& $\vect{f} \ (5.1\text{e}^1)$ 
			& $1.3^{\text{a}}$
			& $46.6^{\text{d}}$
			& $19.0^{\text{a}}$
			& $33.6^{\text{d}}$
			& $\mathbf{2.5}^{\text{a}}$
			& $\mathbf{8.2}^{\text{d}}$
			& $3.2^{\text{a}}$
			& $\mathbf{8.2}^{\text{d}}$ \\ 
			\bottomrule
			\addlinespace[2pt]
			\multicolumn{13}{l}{ \scriptsize $^{\text{a}}$ Total angle between $\e{[tf]}{}{x}$ and $\e{[tf^{*}]}{}{x}$ ~~~~~~~~
				$^{\text{b}}$ Magnitude of common normal between $\e{[tf]}{}{x}$ and $\e{[tf^{*}]}{}{x}$ ~~~~~~~~
				$^{\text{c}}$ Total angle between $\e{[tf]}{}{z}$ and $\e{[tf^{*}]}{}{z}$} \\
			\multicolumn{13}{l}{ \scriptsize $^{\text{d}}$ $\norm{[p^{*}_{x} ~ p^{*}_{y} ~ p^{*}_{z}]^{T} - [p_{x} ~ p_{y} ~ p_{z}]^{T}}$ ~~~~~~~~~~~~~~~~
				$^{\text{e}}$ $\norm{[p^{*}_{x} ~ p^{*}_{y}]^{T} - [p_{x} ~ p_{y}]^{T}}$ }
		\end{tabular}
	}
\end{table*} 
\textit{Viewpoint.}
Viewpoint $\mathcal{TL}$ was consistently selected for the origin in all applications. The decision was particularly significant in all applications where the point of contact moved w.r.t. the world, for example in the Drawing task where the contact force is always applied at the tip of the tool, or which were demonstrated in different positions w.r.t. the world, for example in the Bottle Opening. Conversely, in applications such as Peg on Hole Alignment and Prismatic Joint, where the contact point does not move a lot w.r.t. the world, and/or which were always demonstrated in the same position w.r.t. the world, the decision had a low significance. When analyzing the individual trials of the Peg on Hole Alignment we even observed that $\mathcal{TL}$ was favored for four trials while $\mathcal{W}$ was favored for one trial. Note that we can also consider the hole to be attached to the gripper and the peg to the world, which would result in swapping the viewpoints between $\mathcal{TL}$  and $\mathcal{W}$.

The results for the orientation viewpoint are also plausible. For example, the significant choice for $\mathcal{TL}$ for the Revolute Joint follows from the variability in the demonstrations, with the joint in different orientations w.r.t. the world. The 3-D Contour Following task also exhibits a high ratio, attributed to the low variation in direction in $\mathcal{TL}$ for both force and translational velocity, whereas these directions vary a lot in $\mathcal{W}$. In contrast, the ratios for the Peg on Hole Alignment, 2-D Contour Following and Prismatic Joint indicate only slight preferences for $\mathcal{TL}$ or $\mathcal{W}$. This is in agreement with the limited variability of the orientation in the demonstrations.

\textit{Vectors of Interest.}
Overall the procedure yields plausible results for the vectors of interest.
For motion, Revolute Joint and Peg on Hole Alignment have significant ratios, indicating predominant rotation according to Model 1. Conversely, 3-D Contour Following and Prismatic Joint exhibit high ratios, indicating predominant translation according to Model 2. For 2-D Contour Following, the ratio is closer to 1, indicating a smaller preference for translation over rotation. This is attributed to the tool maintaining a constant orientation while translating in two d.o.f., hence there is not a single predominant direction for translation in both $\mathcal{TL}$ and $\mathcal{W}$.

Concerning wrenches, Model 1 yields the most significant fit for 3-D Contour Following, because the force has quite a constant application point in $\mathcal{TL}$. Remarkably, Bottle Opening is the only task where Model 2 (constant moment) yields the best fit. Accordingly, the moment is selected as the wrench vector of interest.

\begin{figure}[!t]
	\centering
	\includegraphics[height= 2.45cm, page = 1]{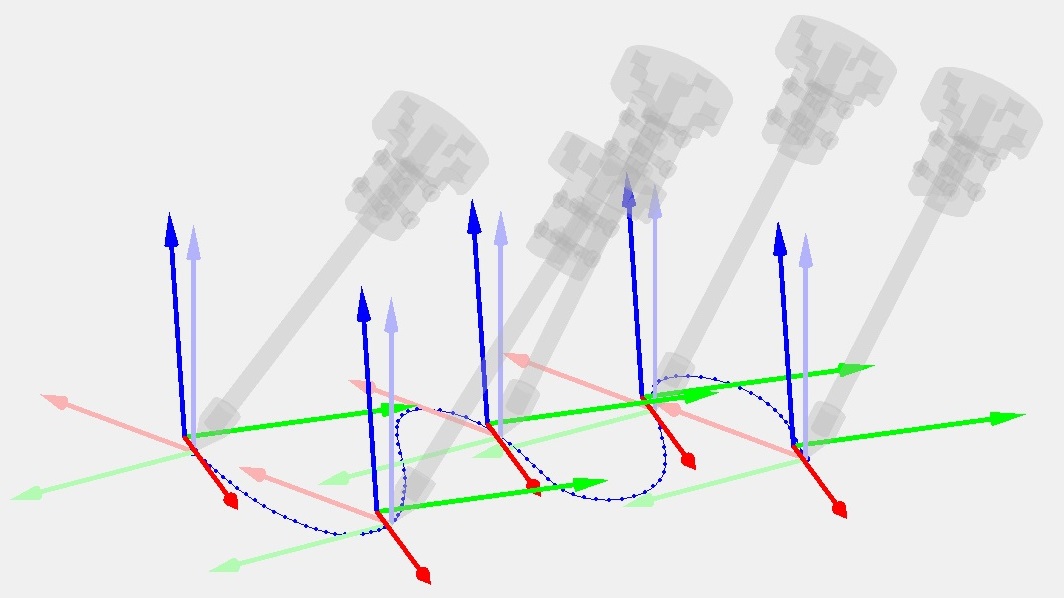}
	\caption{\color{black}Expert task frame (pale) and derived task frame (bright) for the Drawing task according to the RGB convention. The origin of the task frame is attached to the tool while the orientation of the task frame is attached to the world. The derived task frames for the other applications are shown in videos in the supplementary material\color{black}.}
	\label{fig:task_frames}
\end{figure}

\textit{Comparison with Expert's Choice.}
Some, but certainly not all, of the task frame candidates derived from only motion or only wrench data closely align with the expert task frame. For example, the motion candidate for Revolute Joint and the wrench candidate for 2-D Contour Following closely align with $\refframe{tf^{*}}$, both in origin and orientation. Also, it appears that motion candidates are often more accurate, which is attributed to the operator exhibiting more variability in the wrench space.  All this underscores the importance of averaging the results of motion and wrench candidates, which may involve combining complementary information.
\color{black}
The comparison with the expert for the Drawing task is visualized in Fig.~\ref{fig:task_frames}. In this application, the origin of the task frame is attached to the tool while the orientation of the task frame is attached to the world. For the rest of the tasks, please refer to the videos in the supplementary material for the visualization of the derived and expert task frames.
\color{black}
Finally, our results indicate that the weighted average of motion and wrench orientation candidates does not yield substantial benefit compared to the unweighted average, at least for the experiments that we performed.

\textit{Conclusion.}
The experimental results reveal a favorable correspondence between the derived and expert task frames for all the studied applications. They also reveal that the derived task frame encodes the variability that is present in the demonstrations (or lack thereof). Although often less accurate, depending on the application, the wrench candidate proves valuable in determining the optimum origin and orientation in combination with the motion candidate.
There is not a significant benefit in using a weighted average for determining the task frame orientation. However, we did not invest effort in tuning the four reference values, relying on initial guesses instead.

\subsection{Control Performance}
\label{sec:exp_control_performance}

To assess the controller's performance, the tracking errors for all signals and all robot experiments are reported in Table \ref{tab:results} \color{black}using the RMSE. \color{black}

For every application, several scenarios were considered. Below, we present a brief overview of the results, ordered according to the scenarios introduced in Section~\ref{sec:scenarios}. 

\begin{table}[!t]
	\centering
	\caption{Robot control performance: tracking errors for all reference signals in all applications and scenarios.}
	\label{tab:results}
	\resizebox{\linewidth}{!}{
		\begin{tabular}{@{}lccccccc@{}}
			\toprule
			\multicolumn{1}{c}{\multirow{2}{*}{\textbf{Task}}}
			& \multirow{2}{*}{\textbf{No.}}
			& $\Delta R^{a}$
			& $\Delta p$
			& $\Delta \omega$
			& $\Delta v$
			& $\Delta f$
			& $\Delta m$ \\
			
			&
			& $[{}^{\circ}]$
			& $[\text{mm}]$
			& $[\sfrac{{}^{\circ}}{\text{s}}]$
			& $[\sfrac{\text{mm}}{\text{s}}]$
			& $[\text{N}]$
			& $[\text{Nm}]$ \\ [2pt]
			\toprule
			Revolute Joint\\
			~~~Nominal
			& 1-1
			& $0.3$
			& $8.3$
			& $0.1$
			& $4.8$
			& $4.0$
			& $0.5$  \\
			~~~Speed $\times$ 0.25
			& 1-2
			& $0.3$
			& $8.7$
			& $0.0$
			& $2.2$
			& $1.8$
			& $0.3$ \\ [3pt]
			Peg on Hole Alignment\\
			~~~Nominal
			& 2-1
			& $0.4$
			& $1.6$
			& $0.1$
			& $2.6$
			& $1.3$
			& $0.1$  \\
			~~~Wrench $\times$ 3
			& 2-2
			& $0.2$
			& $2.8$
			& $0.1$
			& $3.9$
			& $2.2$
			& $0.2$  \\
			~~~Peg Diameter $\times$ 0.7
			& 2-3
			& $0.2$
			& $3.9$
			& $0.0$
			& $2.9$
			& $1.3$
			& $0.0$  \\ [3pt]
			Bottle Opening \\
			~~~Nominal
			& 3-1
			& $1.1$
			& $1.6$
			& $3.8$
			& $2.2$
			& $7.4$
			& $0.3$  \\
			~~~Opener Length $\times$ 1.5
			& 3-2
			& $3.3$
			& $2.7$
			& $7.3$
			& $4.1$
			& $6.9$
			& $0.3$  \\ [3pt]
			Prismatic Joint \\
			~~~Nominal
			& 4-1
			& $0.1$
			& $3.8$
			& $0.0$
			& $1.5$
			& $1.9$
			& $0.3$  \\
			~~~Speed $\times$ 4.5
			& 4-2
			& $0.0$
			& $3.9$
			& $0.0$
			& $3.7$
			& $4.4$
			& $0.5$  \\
			~~~Rotated Task Frame
			&4-3
			& $0.1$			
			& $2.9$
			& $0.0$
			& $1.3$
			& $1.6$
			& $0.4$  \\ [3pt]
			Drawing \\
			~~~Nominal
			& 5-1
			& $0.1$
			& $5.5$
			& $1.3$
			& $3.1$
			& $2.1$
			& $0.2$  \\
			~~~Wrench $\times$ 0.5
			& 5-2
			& $0.1$
			& $5.4$
			& $1.2$
			& $3.0$
			& $2.1$
			& $0.1$  \\
			~~~Inclined Table
			& 5-3
			& $0.1$
			& $8.1$
			& $1.2$
			& $4.7$
			& $3.8$
			& $0.2$  \\
			~~~Inclined Table, Iterated
			& 5-4
			& $0.1$
			& $2.9$
			& $1.7$
			& $1.5$
			& $0.6$
			& $0.0$  \\ [3pt]
			2-D Contour Following \\
			~~~Nominal
			& 6-1
			& $0.1$
			& $4.4$
			& $0.6$
			& $2.6$
			& $1.1$
			& $0.2$  \\
			~~~No Pose Constraints$^{b}$
			& 6-2
			& $14.2$
			& $80.8$
			& $2.4$
			& $4.4$
			& $1.8$
			& $0.2$  \\ [3pt]
			3-D Contour Following \\
			~~~Nominal
			& 7-1
			& $3.3$
			& $28.1$
			& $1.4$
			& $2.9$
			& $10.1$
			& $0.2$  \\
			~~~Nominal, Iterated
			& 7-2
			& $1.5$
			& $10.9$
			& $1.5$
			& $0.7$
			& $2.7$
			& $0.2$ \\
			~~~Changed Origin$^{b}$
			& 7-3
			& $3.8$
			& $26.6$
			& $1.7$
			& $2.8$
			& $9.4$
			& $0.2$  \\
			\bottomrule
			\addlinespace[2pt]
			\multicolumn{8}{l}{\scriptsize $^{a}$ The RMSE values for $\Delta R$ are calculated as $\norm{\vect{r}}$, with $\vect{r}$ derived according to \eqref{eq:matrix_logarithm_R}.} \\
			\multicolumn{8}{l}{\scriptsize $^{b}$ Task performance was not successful.}
		\end{tabular}
	}
\end{table}

\subsubsection{Nominal Scenarios}
\label{sec:scenario_nominal}
The first six tasks were performed under conditions including task speed that closely resemble those of the demonstrations, see lines 1-1 to 6-1. They were all successful, with small values for the pose, twist and wrench tracking errors. For Bottle Opening, the force error $\Delta f$ was higher than in the other tasks, almost doubled compared to the (small) average desired force. This may be attributed to the significant desired moment, averaging 1.1~$\text{Nm}$, and the expected variability of the required interaction wrench to open different bottles. 
The last task, 3-D Contour Following (7-1), was demonstrated with a translational velocity of approximately 45~$\sfrac{\text{mm}}{\text{s}}$ and a contact force of 50~$\text{N}$, requiring a lot of skill and training by the operator to keep all five wheels in contact. For the performance by the robot to be successful, the translational velocity had to be reduced to 15~$\sfrac{\text{mm}}{\text{s}}$. The value of the force error $\Delta f$ was large compared to the other tasks; however, it accounted for only 20$\%$ of the average value, 50~$\text{N}$, which was much higher than in the other tasks.
Additionally, the small orientation and position errors in all nominal\footnote{Or, for the case of 3-D Contour Following: Nominal, Iterated} scenarios indicate that the calculation of the desired pose using the similarity transformation is analytically correct.

\begin{figure}[!t]
	\centering
	\subfloat[]{\includegraphics[width = 2.45cm, page = 1]{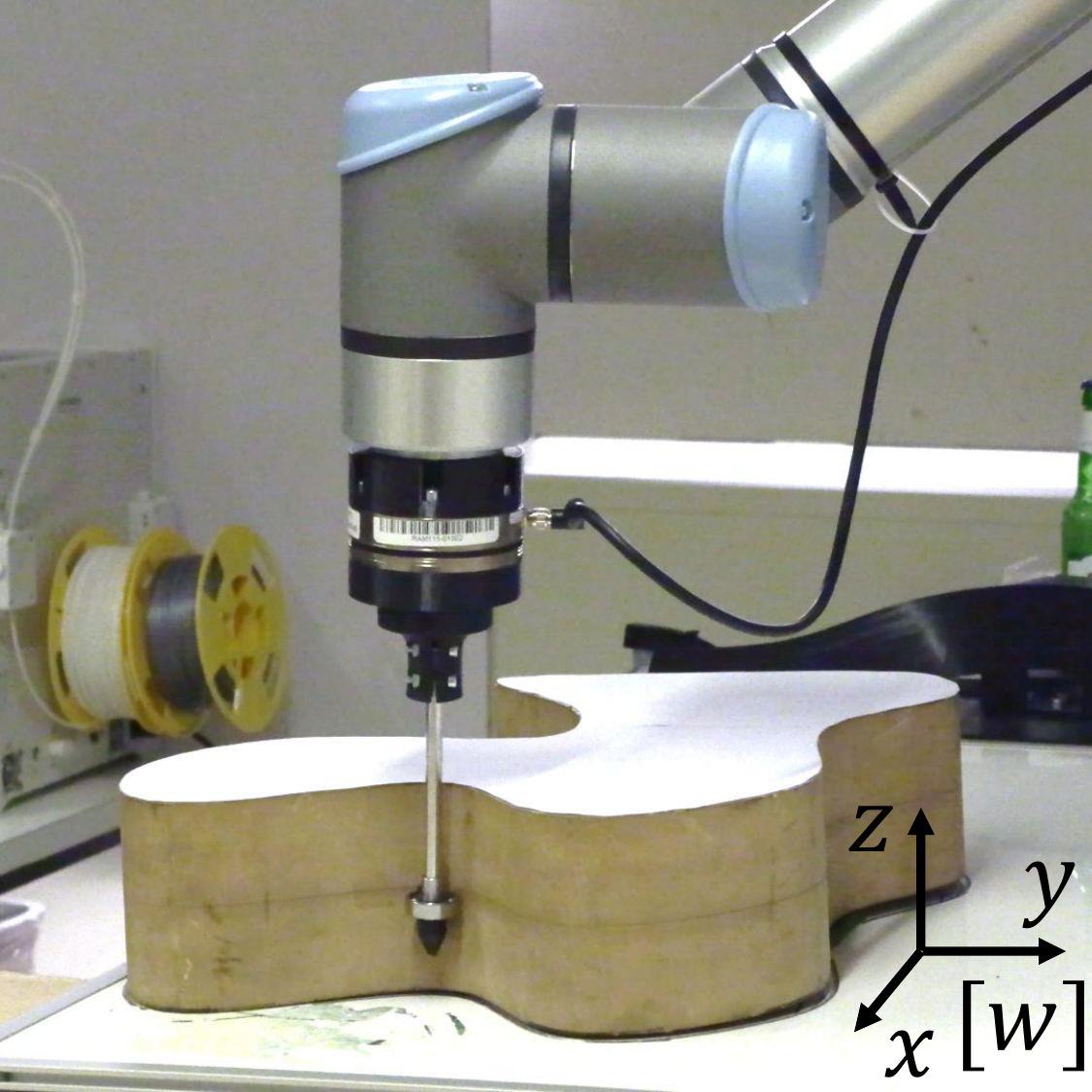}}
	\qquad
	\subfloat[]{\includegraphics[width = 2.45cm, page = 2]{figures/contour2D.pdf}}
	\caption{\color{black} 2-D Contour Following performed by the UR10e robot \color{black} \color{black} (snapshots of video): \color{black} (a) nominal scenario with pose and wrench constraints; (b) nominal scenario without pose constraints, \color{black} resulting in task failure. \color{black}}
	\label{fig:contour2D}
\end{figure}

\textit{Role of pose constraints.}
Pose constraints are required to maintain a constant position or orientation of the tool in the absence of geometric constraints imposed by the environment. This is illustrated in the 2-D Contour Following application (6-2), where we omitted the pose constraints. This led to a failed task, because the orientation of the tool and its vertical position are not geometrically restricted, allowing them to deviate far from their desired values. This is depicted in Fig.~\ref{fig:contour2D} and indicated by the large values of $\Delta R$ and $\Delta p$.

\textit{Role of wrench constraints.}
Controlling the contact wrench allows to compensate for minor disturbances and small deviations from the demonstration environment. This is illustrated in a variation of the Drawing task by tilting the table 5$^{\circ}$ w.r.t. its demonstrated orientation without adjusting the tool's initial orientation (5-3).
The inclined table causes the errors in translational direction,  $\Delta f$, $\Delta p$ and $\Delta v$, to increase w.r.t. the nominal scenario: to maintain contact, force feedback normal to the surface generates a compensating motion.

\begin{figure}[!t]
	\centering
	\subfloat[]{\includegraphics[height = 2.45cm, page = 1]{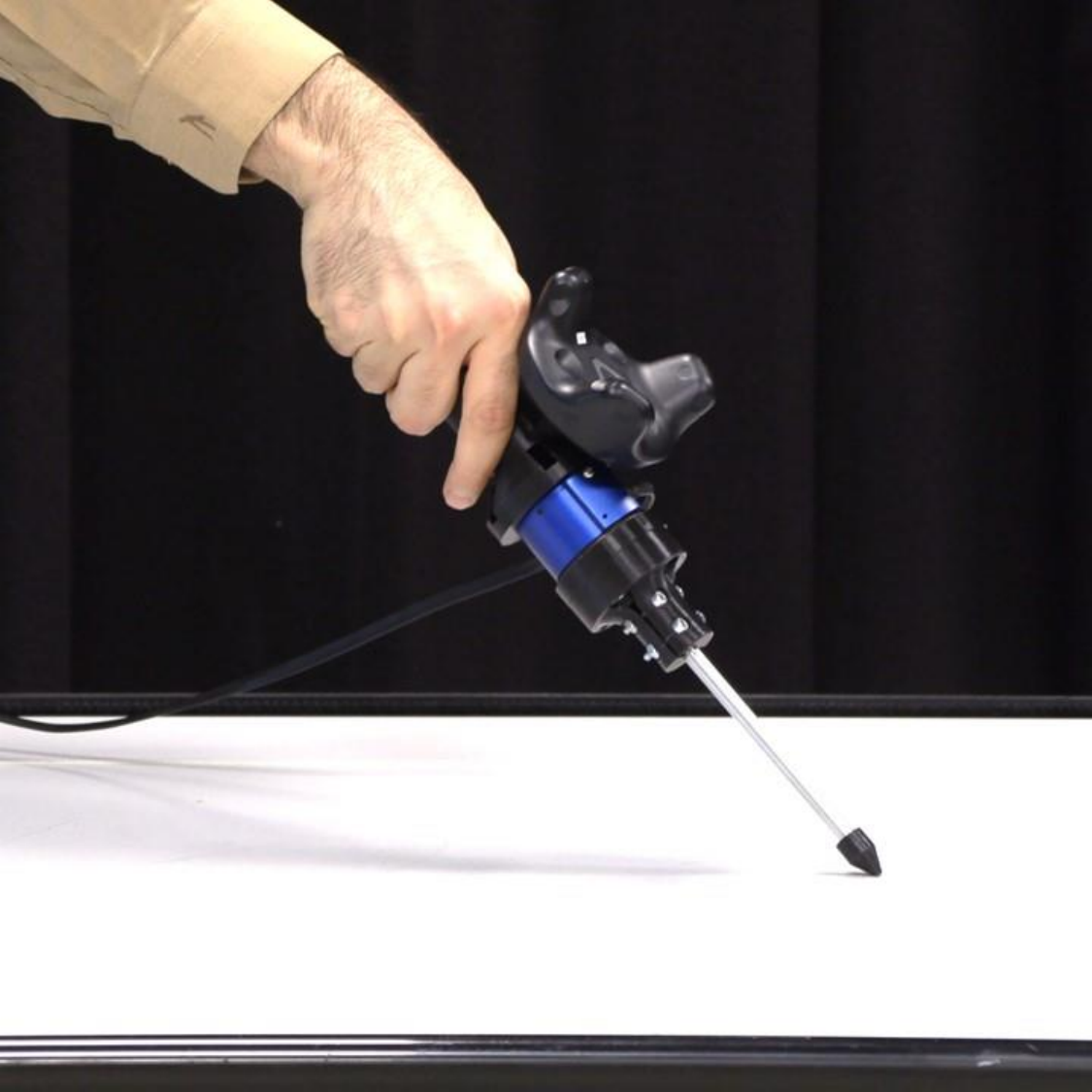}}
	\qquad
	\subfloat[]{\includegraphics[height = 2.45cm, page = 2]{figures/drawing.pdf}}
	\caption{\color{black} Difference in orientation of the tool in the Drawing task \color{black} \color{black} (snapshots of video): \color{black} (a) tilted in the demonstrations; (b) almost vertical in the robot performance}.
	\label{fig:relative_pose}
\end{figure}

\textit{Independence of initial pose.}
This feature is illustrated in the nominal scenario of the Drawing task (5-1), see Fig.~\ref{fig:relative_pose}: the operator demonstrated using a tilted pen, where the initial angle between the normal vector of the surface and the pen was approximately 30$^{\circ}$, and approximately maintained a constant orientation during the task. In contrast, the robot started from an initial pose where the pen was approximately oriented along the normal of the surface, but was able to maintain a similar constant orientation w.r.t. this initial orientation. This is evidenced by the small value of $\Delta R$.

\textit{Variation of task speed.}
Task speed can easily be varied by changing the progress function, $\xi(t)$. An overall speed-up or slow-down is implemented by adjusting $\dot{\xi}$ with a constant factor.
In the Revolute Joint application, the task was performed with 25$\%$ of the demonstrated velocity (1-2). The reduction in speed led to a more precise tracking of the desired values or maintained a comparable level of accuracy compared to the nominal case. This was according to expectation, as lower speeds allow the controller to track the desired values more accurately. The reduction was more pronounced for $\Delta\omega$, $\Delta v$, $\Delta f$ and $\Delta m$.  
Conversely, to assess the controller's capabilities, the Prismatic Joint task was performed at its maximum speed of 180~$\sfrac{\text{mm}}{\text{s}}$, i.e. 4.5 times faster than the nominal case. The results (4-2) indicate a similar level of error for all variables compared to the nominal case, except for $\Delta v$ and $\Delta f$ which increased by a factor up to 2.5. As expected, the desired force could not be tracked as effectively as in the nominal case.

\textit{Variation of contact wrench.}
Similarly, the desired contact wrench can easily be rescaled. 
In the Peg on Hole Alignment application, the desired wrench was tripled (2-2). Accordingly,  $\Delta f$ and  $\Delta m$ almost double in comparison to the nominal case, but they remain small.  Also $\Delta p$ and $\Delta v$ increase but remain small. 
Conversely, the desired wrench values were halved in the Drawing task, still yielding successful completion. The errors are reported in 5-2, consistently demonstrating equal or lower values for all errors compared to the nominal case.

\subsubsection{Changed Geometry}

\begin{figure}[!t]
	\centering
	\subfloat[]{\includegraphics[height = 2.45cm, page = 1]{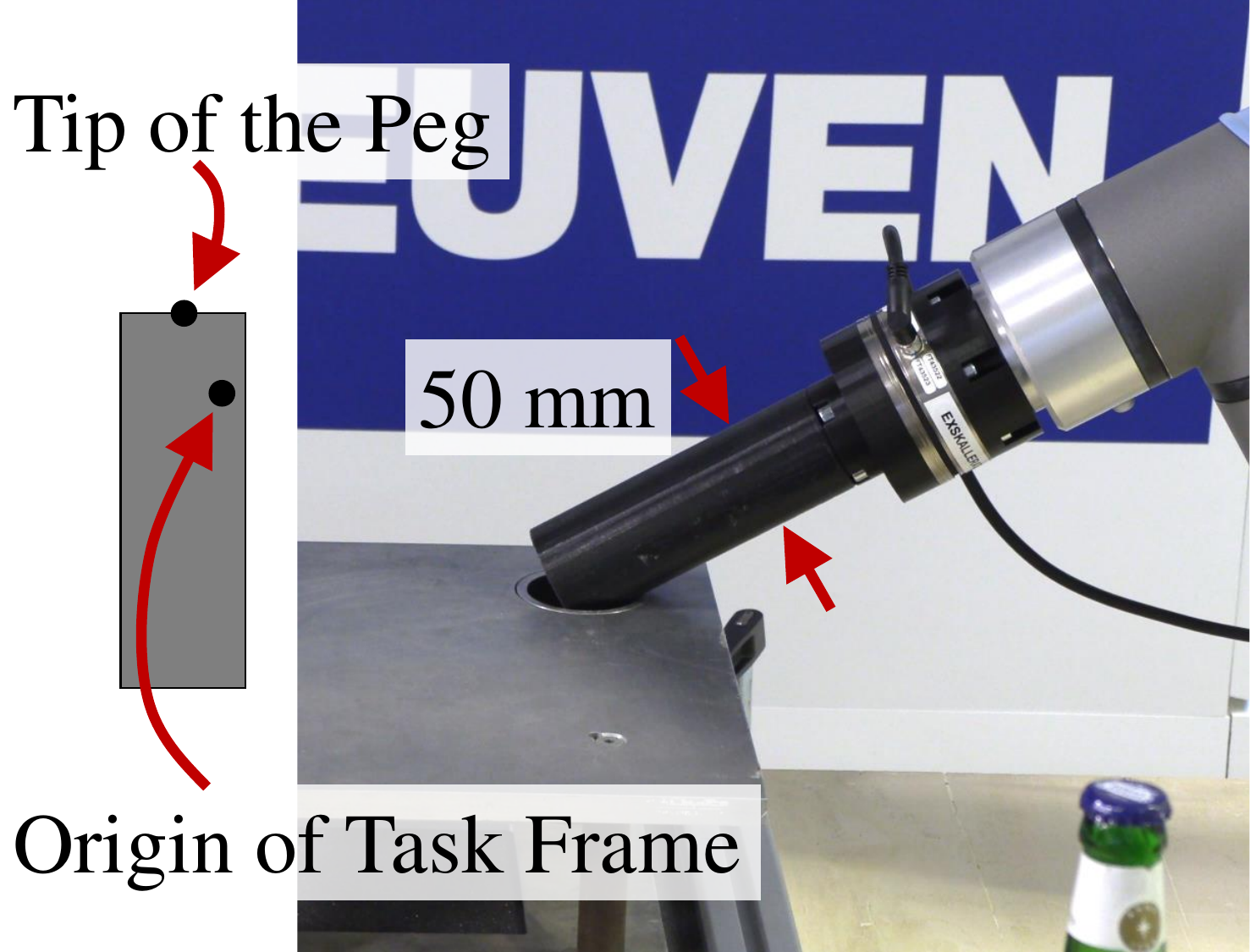}}
	\qquad
	\subfloat[]{\includegraphics[height = 2.45cm, page = 2]{figures/pegs.pdf}}
	\caption{\color{black} Peg on Hole Alignment performed by the UR10e robot \color{black} \color{black} (snapshots of video): \color{black} (a) nominal scenario with peg diameter of 50~$\text{mm}$; (b) changed geometry with peg diameter of 35~$\text{mm}$\color{black}. \color{black}}
	\label{fig:pegs}
\end{figure}

The original peg in the Peg on Hole Alignment task was replaced by a thinner and shorter one, while the hole diameter was reduced accordingly. This is illustrated in Fig.~\ref{fig:pegs}. 
To adapt to this change, we intuitively scaled the position of the task frame relative to a frame at the tip of the new peg in two directions, as illustrated in Fig.~\ref{fig:pegs} to the left of the pictures. The used scaling factor was $\frac{35~\text{mm}}{50~\text{mm}}=0.7$. This adjustment enabled a successful task completion with no significant change in $\Delta f$ and $\Delta m$ compared to the nominal scenario, as reported in line 2-3.

\begin{figure}[!t]
	\centering
	\subfloat[]{\includegraphics[height = 2.45cm, page = 1]{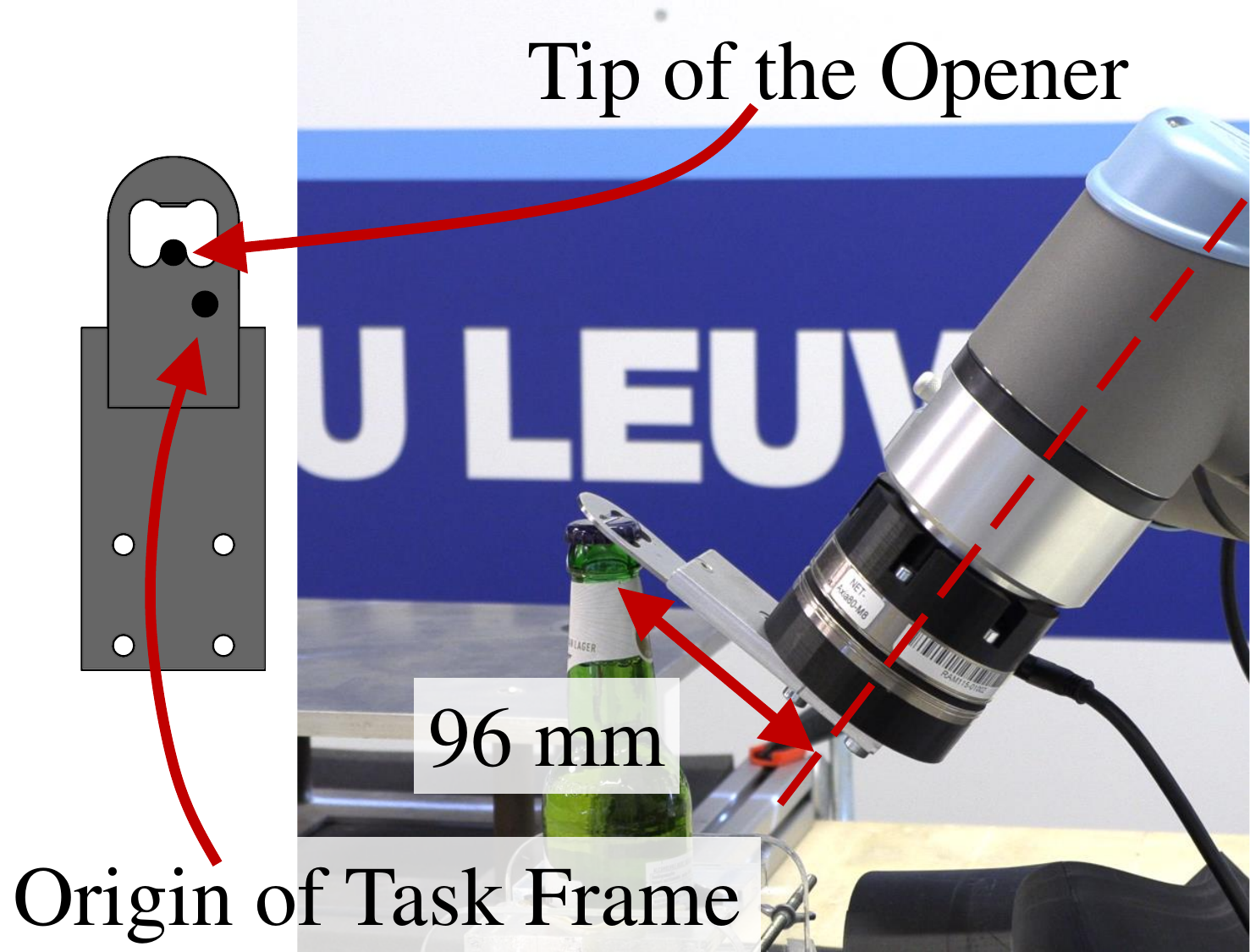}}
	\qquad
	\subfloat[]{\includegraphics[height = 2.45cm, page = 2]{figures/openers.pdf}}
	\caption{\color{black} Bottle Opening performed by the UR10e \color{black} \color{black} (snapshots of video): \color{black} (a) nominal scenario with same opener as in demonstrations; (b) changed geometry with a $50\%$ longer opener\color{black}. \color{black}}
	\label{fig:openers}
\end{figure}

Similarly, in the Bottle Opening application, the opener was replaced by another one which was 
1.5 times longer. Again, we relied on intuition to relocate the task frame, as shown in Fig.~\ref{fig:openers}: the origin of the task frame remains unchanged w.r.t. to a frame attached to the lower inner tip of the opener. Due to the increased length of the opener, the compliance values $\hat{C}_{f}$ and $\hat{C}_{m}$ had to be multiplied by a factor of two. The task frame adjustment enabled a successful cap removal. However, as indicated in 3-2, compared to the nominal scenario we observed similar RMSE levels for $\Delta f$ and $\Delta m$, but nearly doubled RMSEs for the remaining variables.

We could have followed a similar approach in the Drawing scenario with inclined table, i.e. adapt $\R{tf}{w}{}$ manually. Instead, illustrating another approach, we recorded the task performance by the robot with the inclined table (5-3), and reintroduced these data in the task frame derivation procedure to find the change in task frame orientation. Subsequently, we repeated the task performance with inclined table and with the new task frame (5-4). This resulted in smaller errors for almost all variables, even compared to the nominal case. This is not unexpected, because, using robot data instead of human demonstration data avoids calibration errors between demonstration and robot environment. This is an interesting finding. In a similar manner, we derived the task frame from robot data recorded during nominal 3-D Contour Following task and we repeated the task performance. The RMSEs decreased as shown in (7-2), especially for $\Delta{p}$ and $\Delta{f}$. Note that calibration errors may also result from
differences in compliance between the demonstration and robot performance setups, affecting the desired pose trajectory. This was particularly the case in the 3-D Contour Following application, as we introduced some compliance in the robot setup by putting the contour on a foam sheet. The smaller errors in the iterated robot performance allowed us to increase the task speed to 20-25 $\sfrac{\text{mm}}{\text{s}}$ while keeping all five wheels in contact.

\subsubsection{Sensitivity to Task Frame Origin and Orientation}
To confirm that the proposed controller is not dependent on the orientation of the task frame, we repeated the Prismatic Joint task with a 17.7$^{\circ}$ change in orientation for the task frame. As reported in 4-1 and 4-3, the error values remained at the same level, affirming the controller's robustness to variations in the task frame orientation.
Furthermore, to illustrate the relevance of selecting an appropriate origin for the task frame, we changed the origin of the derived task frame by 23.4~$\text{mm}$ in the 3-D Contour Following application. This resulted in a failed task, as some of the wheels lost contact. It is however challenging to confirm loss of contact at some wheels from the reported errors, as the robot kept on moving with some wheels out of contact, but close to the contour. However, the slightly higher values of $\Delta R$ and $\Delta \omega$ are an indication that the tool experienced more rotation than in the nominal case, turning away from the contour.

\section{Discussion and Conclusion}
\label{sec:conclusion}
\color{black}\textit{Contributions.} \color{black}
The main contribution of this article is an automatic procedure for deriving an optimal task frame for contact-rich robot tasks from demonstration data. \color{black} The procedure is summarized in Fig.~\ref{fig:alg_comp} and fits in the red block of the overall workflow depicted in Fig.~\ref{fig:overview_b}. The demonstration data may consist of human-demonstrated, simulation or robot data. \color{black}The procedure is based on two principles that are hypothesized to 
underpin the control configuration targeted by an expert when choosing the task frame: (1) decoupled control of the directional and moment components of a screw, and (2) decoupled control of motion and interaction wrench. The procedure is rooted in screw theory and consists of a pipeline of algorithms. Besides existing algorithms such as ASIP and AVOF, the pipeline includes the following novelties: (1) an alternative application of ASIP (i.e. ASIP applied to a mean-subtracted screw), (2) a new algorithm to align orientation frames, (3) an adapted version of an algorithm to find the average of two orientations that can be applied to rotation matrices instead of quaternions, and (4) an extension of the concept of similarity transformation to pose matrices. The procedure is entirely probabilistic and does not involve any hyperparameters unless optional weighting is applied for averaging the orientation (but in our experiments we did not find evidence of significant added value for the latter). It also produces invariant results, regardless of the reference frames in which the data are expressed. The result is a gray-box model of the task, i.e. a set of parameters and reference signals that have unambiguous physical meaning and relations between them, and that are obtained in a data-driven way.

\color{black}\textit{Main innovations with respect to the state of the art.}
The task frame derivation approach is fully automatic and fully data-driven. I.e., it is \emph{generative}, as it does not require predefined candidate reference frames or any input other than the demonstration data. Furthermore, the task frame is not constrained to be entirely fixed to the world or the tool as the origin and the orientation may be - independently - fixed to either world or tool.
The approach \color{black}systematically uses (1) both motion and wrench data and (2) complete screw data, i.e. both in translation and orientation, and force and moment. \color{black}As a result, the approach is \emph{generic}, as it applies to a large variety of contact-rich tasks.\color{black}

\color{black}\textit{Validation of the task frame derivation procedure.}
First\color{black}, the task frame derivation procedure \color{black}itself \color{black}was validated using seven applications including surface following tasks (2-D and 3-D Contour Following, Drawing), manipulation of articulated objects (Revolute and Prismatic joints),  a nonreciprocal task involving plastic deformation (Bottle Opening) and tasks involving multiple contact points (Peg on Hole Alignment, 3-D Contour Following). The tasks involved both activities of daily life and technical tasks. While the derived task frames showed a good agreement with the task frames assumed to be chosen by experts, it is important to note that, (1) depending on the type of application, the accuracy of all the components of the origin or orientation may not be equally relevant, (2) in some cases the derived frame may be more optimal than the expert’s choice, which may be based on intuition or assumptions that are too simplified. Also, although the experiments were based on five demonstration trials for each application, the procedure can be applied to a single demonstration as well. Finally, our experiments confirmed the added value of combining motion and wrench data for deriving the optimal task frame.

\color{black}\textit{Validation of the benefits of using a well-chosen and parameterized task frame model.} \color{black}A second set of validation experiments consisted of a robot performing the seven tasks based on the obtained task model (task frame parameters and reference signals for twist, wrench and pose). We designed a one-size-fits-all controller based on the combination of pose and wrench constraints. Hence, the controller was certainly not optimal for all applications – that was not the main focus of the article – but could be tuned quickly for each application with only three parameters. The experiments showed that, for each application, successful task completion and stable control could be achieved in the nominal scenario where the robot performs the task under the same conditions as observed during the demonstrations. Furthermore, thanks to the type of controller and the gray-box nature of the task model, we could demonstrate successful task performances in case of (1) a different starting pose, (2) a changed task speed, (3) a changed magnitude of the interaction wrench and (4) small changes to the geometric configuration. For larger changes in the geometric configuration of tool or environment, we demonstrated two solutions: (1) manual adaptation of the task frame parameters and (2) iteration of the derivation of the task frame parameters using the data of the task performed by the robot. The latter option is interesting because such iteration eliminates calibration errors between the demonstration environment and the robot environment. Further investigation is required, but this option could potentially turn out to be a good standard practice after an application has been ported from demonstration to robot. Finally, it was shown that, due to its isotropic nature, the controller performed equally well regardless of the orientation of the task frame, while its performance deteriorated when the origin of the derived task frame was changed. The latter underscores the importance of selecting the optimal origin for the task frame.

\color{black}\textit{Limitations and future work.}
Current limitations of the proposed method are the following. \color{black} First, we started from data that had already been segmented. Future work could investigate whether changes in the outcomes of the task frame derivation procedure can be used to identify transitions between the contact-rich segments studied in this article and other types of segments: gross motions in free space,  approach-to-contact motions, building up of an initial contact and moving out of contact.

Second, the task frame derivation procedure includes several binary decisions (e.g. about viewpoint or progress variable). The choice between the two options can sometimes be very arbitrary, just depending on noise and small human variations present in the data. However, we argue that, in such cases, both options yield a  task model that will result in comparable robot task performance. Nevertheless,  care has to be taken when the decision is used for other purposes, such as segmentation (see above). Therefore it is important to always consider also the statistical significance of the decision (i.e. the ratio of the determinants of the covariance matrices).

Third, we used reference signals obtained by only averaging the demonstration signals of the different trials after aligning them. We did not use more advanced statistics to deal with noise or with undesired human variations in the demonstration data. Also, we did not estimate the impedance observed in the human demonstrations and employ it in the robot controller, which is known to come at the cost of a more complicated experimental setup [4]. This was left for future work.

Fourth, we used a one-size-fits-all isotropic type of controller that does not exploit knowledge about contact dynamics or human impedance in different directions. Hence, although our work was inspired by hybrid control, we did not use a hybrid controller. While our approach is simple, it is expected that non-isotropic controllers can further improve the task performance. This is recommended for future work.

Fifth, we derived an \textit{average} task frame for the \textit{whole segment}, with a \textit{fixed} origin and a \textit{fixed} orientation, both w.r.t. either world or tool. A future extension would be to derive an optimal task frame without these restrictions, i.e. an \textit{instantaneous} task frame that is \textit{completely free} to move. Such frame would be beneficial for controlling tasks with changing contact geometries, as in 2-D and 3-D Contour Following involving unknown objects or cutting. \color{black}This could be achieved by considering the average task frame as proposed in this paper, but over a shorter time window, or by switching to invariant motion and wrench trajectory descriptors as in \cite{vochten2023invariant}.\color{black} 

Lastly, although the Bottle Opening application involved plastic deformation of the bottle cap, all our experiments involved rather stiff contacts. Hence, it would be worthwhile to investigate the suitability of our approach when manipulating more deformable objects.


\appendices

\section{Combining Orientation Estimates}
\label{sec:appendix_combining}

\newcommand{\stochvect}[1]{\hat{\boldsymbol{#1}}} 
\newcommand{\stochmat}[1]{\hat{\boldsymbol{#1}}} 
\newcommand{\maxl}[1]{{#1}^{*}} 
\newcommand{\maxlit}[2]{{{#1}^{(#2)}}} 

This appendix explains the semantics of the covariance matrix on the orientation as used in this text and discusses Algorithm~\ref{alg:wei_ave_orien} that combines two orientations estimates $\mat{R}_1$ with covariance matrix $\mat{C}_1$ and $\mat{R}_2$ with covariance $\mat{C}_2$ into a new estimate $\mat{R}$ with covariance $\mat{C}$.

A probabilistic orientation estimate $\stochmat{R}_i$ can be described by an orientation $\mat{R}_i$ and a covariance $\mat{C}_i$ on a probabilistic vector $\stochvect{\delta}$ using
$\stochmat{R}_i = \operatorname{exp}( [\stochvect{\delta} ]_\times ) \mat{R}_i$ with 
$\stochvect{\delta} \sim  \mathcal{N}(\vect{0}, \mat{C}_i)$. 
Therefore, $\stochmat{R}_i$ represents the following distribution:
\begin{align}
	\stochmat{R}_i & \sim P(\mat{R}| \mat{R}_i,\mat{C}_i) = 
	\lambda_i
	\operatorname{exp}\left( -\tfrac{1}{2} \vect{\delta}_i^T \mat{C}_i^{-1} \vect{\delta}_i\right),
\end{align}
with
\begin{align}
	\vect{\delta}_i  = \operatorname{log}^\vee \Big( \mat{R}  \mat{R}_i^{-1} \Big) 
	\text{ and }
	\lambda_i =  (2\pi )^{-3/2}\det({\mat{C}_i })^{-1/2}
	.
	\label{eq:rot_delta_solve}
\end{align}

The maximum likelihood estimate $\maxl{\mat{R}}$ given two estimates $\stochmat{R}_1$ and $\stochmat{R}_2$ described by  $\mat{R}_1$, $\mat{C}_1$, $\mat{R}_2$ and $\mat{\mat{C}}_2$ is:
\begin{align}
	\maxl{\mat{R}}= \max_{\mat{R}}  ~~ \lambda_1 \lambda_2 \operatorname{exp} \left( -\tfrac{1}{2} \vect{\delta}_1^T \mat{C}_1^{-1} \vect{\delta}_1 \right)
	\operatorname{exp}  \left( -\tfrac{1}{2} \vect{\delta}_2^T \mat{C}_2^{-1} \vect{\delta}_2 \right) \nonumber
\end{align}
Noting that the logarithm is a monotonic function, the above optimization is equivalent to the following minimization:
\begin{align}
	\maxl{\mat{R}} & = \min_{\mathbf{R}}  ~~ \tfrac{1}{2} \vect{\delta}_1^T \mat{C}_1^{-1} \vect{\delta}_1  +
	\tfrac{1}{2}  \vect{\delta}_2^T \mat{C}_2^{-1} \vect{\delta}_2 , \label{eq:optim_rot}
\end{align}
This is a difficult minimization problem due to the non-linear relationships between $\mat{R}$ and $\vect{\delta}_i$ in \eqref{eq:rot_delta_solve}:
\begin{align}
	\maxl{\mat{R}} & = \min_{\mat{R}}  ~~ \tfrac{1}{2}  \operatorname{log}^\vee \left( \mat{R} \mat{R}_1^{-1}  \right)^T \mat{C}_1^{-1} 	\operatorname{log}^\vee \left( \mat{R} \mat{R}_1^{-1}  \right)  + \nonumber \\
	&~~~~~~~~~~~  \tfrac{1}{2}  \operatorname{log}^\vee \left( \mat{R} \mat{R}_2^{-1}  \right)^T \mat{C}_2^{-1} 	\operatorname{log}^\vee \left( \mat{R} \mat{R}_2^{-1}  \right)   
	\label{eq:optim_rot_expanded}
\end{align}
Therefore, a numerical procedure is designed according to the same principles as \cite{kavan2006dual}, where we iteratively express all variables with respect to an approximation $\maxlit{\mat{R}}{k}$ of the maximum likelihood $\maxl{\mat{R}}$; then approximate the resulting optimization problem as a quadratic optimization and solve this optimization; after which we update our approximation of $\maxl{\mat{R}}$ to $\maxlit{\mat{R}}{k+1}$ and repeat the procedure.

In iteration step $k$, expressing the variables in \eqref{eq:optim_rot} with respect to $\maxlit{\mat{R}}{k}$ gives:
\begin{align}
	\mat{R} &= \operatorname{exp}\left( [ \vect{r}^{(k)} ]_\times\right) \maxlit{\mat{R}}{k} ,&
	\mat{R}_i &= \operatorname{exp}\left( [ \vect{r}_i^{(k)} ]_\times\right) \maxlit{\mat{R}}{k} ,	
	\label{eq:def_r}
\end{align}
for $i=1, 2$.  When $\vect{r}^{(k)}$ is small, we can approximate $\mat{R}\mat{R}_i^{-1} = \exp([\vect{r}^{(k)}]_{\times}) \exp(-[\vect{r}_i^{(k)}]_{\times})$ as $\exp([\vect{r}^{(k)}-\vect{r}_i^{(k)}]_{\times})$. Using this approximation, \eqref{eq:optim_rot_expanded} reduces to a quadratic optimization problem that we solve to obtain a better estimate $\maxl{\mat{R}}$:
\begin{align}
	\maxlit{\vect{r}}{k+1} =  \min_{\vect{r}^{(k)}}  ~~ \frac{1}{2} (\vect{r}^{(k)} -\vect{r}_1^{(k)})^T \mat{C}_1^{-1} (\vect{r}^{(k)}-\vect{r_1}^{(k)}) + \nonumber \\
	\frac{1}{2} (\vect{r}^{(k)} - \vect{r}_2^{(k)})^T \mat{C}_2^{-1} (\vect{r}^{(k)} - \vect{r_2}^{(k)}),~ \label{eq:optim_rot2}
\end{align}
$\vect{r}_i^{(k)}$ is defined by \eqref{eq:def_r} and computed using $\log^\vee ( \mat{R}_i \maxlit{\mat{R}}{k}^{-1} )$.
The solution of this quadratic optimization problem is:
\begin{align}
	\maxlit{\vect{r}}{k+1} =& 
	~\mat{C} 
	\left( 
	\mat{C}_1^{-1} \vect{r}_1^{(k)}  + \mat{C}_2^{-1} \vect{r}_2^{(k)}
	\right).	 
	\label{eq:compute_delta}
\end{align}
A better approximation $\maxlit{\mat{R}}{k+1}$ of the likelihood is then computed by:
\begin{align}
	\maxlit{\mat{R}}{k+1} &= \exp \left( [ \maxlit{\vect{r}}{k+1} ]_\times\right) \maxlit{\mat{R}}{k}, 
\end{align}
after which the complete procedure can be repeated to further improve the estimate until the updates $r^{(k)}$ become sufficiently small.

The covariance matrix $\mat{C}$ on our estimates $\maxlit{\vect{r}}{k+1}$ is not influenced by the above iterations and is given by:
\begin{align}
	\mat{C}^{-1} = \mat{C}_1^{-1} + \mat{C}_2^{-1}
	\label{eq:combining_covariance}
\end{align}

The complete procedure is found in Algorithm~\ref{alg:wei_ave_orien}.  The above algorithm was first proposed for quaternions and dual quaternions in \cite{kavan2006dual} but was adapted here to rotation matrices. Our covariance matrices are also expressed in the tool or world frame instead of the object frame.

The algorithm results in a unique solution in a few iterations, unless there is a 180$^{\circ}$ difference between the orientations $\mat{C}_1$ and $\mat{C}_2$.  The latter cannot occur since the rotation matrices were aligned using our Algorithm~\ref{alg:permutation}.

\bibliographystyle{IEEEtran}
\bibliography{bibfile}


\end{document}